\documentclass[twoside,letterpaper]{article}

\usepackage[preprint]{aistats2027}
\usepackage[round]{natbib}

\usepackage[hyphens]{url}
\usepackage{graphicx}
\usepackage{booktabs}
\usepackage{float}
\usepackage{stfloats}
\usepackage{algorithm}
\usepackage{algorithmic}
\usepackage{caption}
\usepackage{pdflscape}
\usepackage{placeins}
\usepackage{needspace}

\usepackage{amsmath,amssymb,amsfonts}

\newcommand{\N}{\mathcal{N}}
\newcommand{\KL}{\mathrm{KL}}
\newcommand{\tr}{\mathrm{tr}}
\newcommand{\E}{\mathbb{E}}

\newcommand{\Dff}{\mathbf{D}_{\mathbf{ff}}}
\newcommand{\Kff}{\mathbf{K}_{\mathbf{ff}}}
\newcommand{\Qff}{\mathbf{Q}_{\mathbf{ff}}}
\newcommand{\Kuu}{\mathbf{K}_{\mathbf{uu}}}
\newcommand{\Kfu}{\mathbf{K}_{\mathbf{fu}}}
\newcommand{\Kuf}{\mathbf{K}_{\mathbf{uf}}}
\newcommand{\Id}{\mathbf{I}}
\newtheorem{theorem}{Theorem}
\newtheorem{lemma}{Lemma}
\newtheorem{proposition}{Proposition}

\usepackage{xcolor}

\usepackage[colorlinks=true,linkcolor=blue!50!black,citecolor=blue!50!black,urlcolor=blue!50!black]{hyperref}

\begin{document}

\runningtitle{Geometry-Aware Block Selection for Sparse Variational GPs}
\runningauthor{Matteo Hoxha, Daniel Paulin}

\twocolumn[
\aistatstitle{Geometry-Aware Block Selection for\\Sparse Variational Gaussian Processes}

\aistatsauthor{Matteo Hoxha \And Daniel Paulin}

\aistatsaddress{Nanyang Technological University, Singapore\\\texttt{MATTEO001@e.ntu.edu.sg} \And Nanyang Technological University, Singapore\\\texttt{daniel.paulin@ntu.edu.sg}}
]

\begin{abstract}
Sparse variational Gaussian processes (SVGPs) are a widely used, principled approach to scalable Gaussian process modelling. Their success stems from their theoretically interpretable structure: they provide a lower bound on the marginal likelihood of the full GP while allowing scalable stochastic optimization.
Recent work has explored modifications of the posterior conditional distribution and has shown that such extensions can improve upon this well-established framework. In particular,
introducing a block-diagonal structured
scaling matrix leads to a tighter lower bound
on the GP marginal likelihood.
We investigate whether selecting blocks that tighten the variational objective improves test RMSE and NLL at a fixed inducing-point budget.
Since the scaling matrix contributes to the variational lower bound through an upper bound on the log-determinant of the full residual covariance matrix, our work makes the following contributions. First, we introduce a simple, scalable, geometry-aware algorithm, with explicit computational complexity and low computational overhead, that recovers a large fraction of this log-determinant contribution by selecting an appropriate partition of the data. Second, we show how to train these models using a doubly stochastic optimization framework. Finally, we investigate how sparse GP methods based on alternative conditional posterior structures can be used to learn better hyperparameters and inducing points for SVGP.

\end{abstract}

\section{Introduction}
\label{sec:intro}

Gaussian processes (GPs) are a widely used framework for modelling functions in machine learning and statistical learning. Given training inputs \(\mathbf X=\{x_i\}_{i=1}^N\) and observed responses \(\mathbf y=(y_1,\ldots,y_N)^\top\), they provide a Bayesian treatment of uncertainty and control model complexity through the prior \citep{rasmussen2006}. However, exact GP regression requires computations involving a dense \(N\times N\) covariance matrix, and predictions also depend on the training data. This cost has motivated a large body of work on scalable approximations that reduce the computational requirements of training and prediction while retaining the benefits of GPs \citep{quinonero2005,liu2020}. One of the most widely used approaches is sparse variational inference, introduced by \citet{titsias2009} and extended to stochastic optimization by \citet{hensman2013}. Its central idea is to represent the information in \(N\) observations through \(M\ll N\) inducing variables. These are function values at auxiliary inputs, which need not coincide with observed inputs and can be optimized during training. The resulting training cost is \(O(NM^2)\), or \(O(|\mathcal B| M^2+M^3)\) per stochastic update with a likelihood minibatch \(\mathcal B\). Standard SVGP assumes that the variational conditional over function values given the inducing variables matches the prior conditional. This construction yields a lower bound on the GP log marginal likelihood. Recent work by \citet{titsias2025} and \citet{bui2025tight} shows that relaxing this assumption can produce tighter bounds by enriching the variational family.

In particular, \citet{bui2025structured} introduce a block-diagonal scaling of the conditional posterior covariance. For blocks of size \(b\), that is \(B=\lceil N/b\rceil\) blocks, the block factorizations cost \(O(Nb^2)\) over the full dataset, in addition to the inducing-point computations. The resulting sum of block log-determinants provides an upper bound on the dense residual log-determinant and hence a lower bound on its negative contribution to the objective. Since the dense determinant is invariant under simultaneous row and column permutations, the question we address is: \emph{how can we efficiently partition the residual matrix into informative blocks?} Orderings and groupings play a similar role in Vecchia approximations \citep{guinness2018permutation}. We present a simple geometry-aware algorithm with an explicit search cost of \(O(NM+M^2+N\log N)\) kernel-distance work per refresh. Rather than forming equal-size blocks at random, as in the stochastic experiments of \citet{bui2025structured}, we select blocks of the same size using the inducing-point geometry. Our experiments show how this selection can tighten the bound further at the same block budget. We also show that, when the data form compact groups between consecutive inducing points, the algorithm returns these groups, which under a separation condition form the ELBO-optimal partition into blocks of size \(b\) (Theorem~\ref{thm:geometric-recovery}). We also describe a doubly stochastic implementation that samples likelihood minibatches and residual blocks separately, addressing the localized likelihood updates that would otherwise result from using selected blocks for both terms.

Changing the conditional posterior also changes the cost of prediction. The predictive mean keeps its sparse form, but the predictive variance acquires a correction involving the dense \(N\times N\) residual covariance \(\Dff\) of the training inputs. At test time we approximate this correction blockwise, in the manner of the partially independent conditional (PIC) approximation \citep{snelson2007}: each test input keeps its residual cross-covariance with a single block of training points (Section~\ref{sec:bounds}). This is a prediction-time approximation, not the exact variational covariance. Previous work avoids the correction by replacing the variational conditional with the prior conditional at test time, \(q(f\mid\mathbf{u})\approx p(f\mid\mathbf{u})\) \citep{bui2025tight,bui2025structured}. We interpret this approximation as transferring the learned state from BT-SVGP to a standard SVGP predictor, and study which benefits of block-structured training survive subsequent sparse refinement.

Our code reproducing all experiments in the paper is available at \url{https://github.com/Jollikop/geometry-aware-block-selection}.

\section{Sparse Bounds}
\label{sec:bounds}
We review sparse variational Gaussian processes
\citep{titsias2009,hensman2013,hensman2015} and the recent improvements to
their bounds provided by T-SVGP \citep{titsias2025,bui2025tight} and
BT-SVGP \citep{bui2025structured}. We consider supervised regression on the
training inputs \(\mathbf X=\{x_i\}_{i=1}^N\) and responses
\(\mathbf y=(y_1,\ldots,y_N)^\top\). We seek a function \(f\)
relating inputs to observations, place the GP prior
\(p(f\mid\gamma)=\mathrm{GP}(f;0,k_\gamma)\) on it, and use the Gaussian
likelihood
\(p(\mathbf{y}\mid\mathbf{f},\sigma^2)=\N(\mathbf{y};\mathbf{f},\sigma^2\Id_N)\),
where \(\mathbf{f}=f(\mathbf{X})\).
The kernel and likelihood
hyperparameters are \(\gamma\) and \(\sigma^2\), respectively; we use
\(\theta=(\gamma,\sigma^2)\) to denote them jointly, and write \(k_\theta\) for \(k_\gamma\). Exact evaluation of the
log marginal likelihood requires \(O(N^3)\) time and \(O(N^2)\) memory,
which limits scalability \citep{rasmussen2006}. Sparse variational inference
introduces inducing points
\(\mathbf{Z}=\{z_m\}_{m=1}^M\), with \(M\ll N\), and inducing variables
\(\mathbf{u}=f(\mathbf{Z})\).

Define \(\Qff=\Kfu\Kuu^{-1}\Kuf\) and the prior conditional covariance
\(\Dff=\Kff-\Qff\). We assume \(\Kuu\succ0\); the Schur complement
then gives \(\Dff\succeq0\). We consider the variational model
\citep{bui2025structured}
\[
\begin{aligned}
q(f)&=p(f_{\neq\mathbf{f},\mathbf{u}}\mid\mathbf{f},\mathbf{u})
       q(\mathbf{f}\mid\mathbf{u})q(\mathbf{u}),\\
q(\mathbf{f}\mid\mathbf{u})
    &=\N(\mathbf{f};\Kfu\Kuu^{-1}\mathbf{u},\mathbf{C}).
\end{aligned}
\]
Here \(f\) denotes the full latent function, whereas \(\mathbf{f}\) and
\(\mathbf{u}\) are its values at \(\mathbf{X}\) and \(\mathbf{Z}\).
Following \citet{bui2025structured}, \(f_{\neq\mathbf{f},\mathbf{u}}\)
denotes the remaining function values; their conditional distribution is
kept equal to the prior conditional.
For fixed \(\theta\), \(\mathbf Z\), and \(\mathbf C\), maximizing the
variational lower bound over \(q(\mathbf u)\) gives
\(
q^\star(\mathbf{u})\propto p(\mathbf{u})
\N(\mathbf{y};\Kfu\Kuu^{-1}\mathbf{u},\sigma^2\Id_N).
\)
Substitution yields the collapsed bound
\begin{equation}
\begin{aligned}
\mathcal{F}(\theta,\mathbf{C})
 &=\log\N(\mathbf{y};0,\Qff+\sigma^2\Id_N)\\
 &\quad-\tfrac12\tr[(\sigma^{-2}\Id_N+\Dff^{-1})\mathbf{C}]\\
 &\quad-\tfrac12\log|\mathbf{C}^{-1}\Dff|+\tfrac{N}{2}.
\end{aligned}
\label{eq:general-c-bound}
\end{equation}

The inverse expressions in this derivation are stated for nonsingular
\(\Kuu\), \(\Dff\), and \(\mathbf C\). The optimized log-determinant
bounds below extend to positive-semidefinite \(\Dff\) by continuity.

As shown by \citet{bui2025structured}, different restrictions on
\(\mathbf{C}\) produce different bounds. Without a structural restriction,
the optimum is the dense matrix
\(
\mathbf{C}^\star=(\Dff^{-1}+\sigma^{-2}\Id_N)^{-1},
\)
which gives the dense residual endpoint, denoted E-SGPR,
\begin{equation}
\begin{aligned}
\mathcal{F}_{\mathrm{E\mbox{-}SGPR}}
 &=\log\N(\mathbf{y};0,\Qff+\sigma^2\Id_N)\\
 &\quad-\tfrac12\log|\Id_N+\sigma^{-2}\Dff|.
\end{aligned}
\label{eq:esgpr}
\end{equation}
We use this endpoint in the diagnostic experiments as a reference for
assessing how closely the block method recovers the full residual
log-determinant contribution.

To express the structural choices, write
\(\mathbf{C}=\Dff^{1/2}\mathbf{M}\Dff^{1/2}\), using the symmetric
square root. The matrix \(\mathbf{M}\) is a covariance scaling and is
distinct from the inducing-point count \(M\). For a partition
\(\mathcal{P}=\{\mathcal{B}_r\}_{r=1}^B\) of the observation indices
\([N]=\{1,\ldots,N\}\), choosing
\(\mathbf{M}\) block diagonal and optimizing its blocks gives
\(\mathbf{M}_r^\star=(\Id_{|\mathcal{B}_r|}
+\sigma^{-2}\Dff[\mathcal{B}_r,\mathcal{B}_r])^{-1}\) and
\begin{equation}
\begin{aligned}
\mathcal{F}_{\mathrm{BT}}(\mathcal{P})
 &=\log\N(\mathbf{y};0,\Qff+\sigma^2\Id_N)\\
 &\quad-\tfrac12\sum_{r=1}^B
 \log\left|\Id_{|\mathcal{B}_r|}
 +\sigma^{-2}\Dff[\mathcal{B}_r,\mathcal{B}_r]\right|.
\end{aligned}
\label{eq:bt}
\end{equation}
This is the block tightening of \citet{bui2025structured}. As in their
experiments, we use equal-size blocks. We write (R)BT-SGPR for this bound evaluated
on a uniformly random equal-size partition and
(S)BT-SGPR for the same bound evaluated on the selected partition of
Section~\ref{sec:selection}. The prefixes (R) and (S) stand for random and
selected blocks and apply equally to BT-SVGP. Following their convention, \(B\)
denotes the number of blocks; we write \(b=N/B\) for the block size when
\(B\) divides \(N\), allowing a smaller final block otherwise.
A diagonal \(\mathbf{M}\), one entry per observation, yields T-SGPR
\citep{titsias2025,bui2025tight}:
\begin{equation}
\begin{aligned}
\mathcal{F}_{\mathrm{T}}
 &=\log\N(\mathbf{y};0,\Qff+\sigma^2\Id_N)\\
 &\quad-\tfrac12\sum_{i=1}^N\log\!\left(1+\frac{[\Dff]_{ii}}{\sigma^2}\right).
\end{aligned}
\label{eq:t}
\end{equation}
Finally, fixing \(\mathbf{M}=\Id_N\) recovers the SGPR bound
\citep{titsias2009}:
\begin{equation}
\begin{aligned}
\mathcal{F}_{\mathrm{SGPR}}
 &=\log\N(\mathbf{y};0,\Qff+\sigma^2\Id_N)\\
 &\quad-\tfrac{1}{2\sigma^2}\tr(\Dff).
\end{aligned}
\label{eq:sgpr}
\end{equation}
For fixed hyperparameters and inducing points, the Hadamard--Fischer
inequality and \(\log(1+x)\leq x\) imply the hierarchy
\citep{bui2025structured,braun2023hfk}
\[
\begin{aligned}
\mathcal{F}_{\mathrm{SGPR}}
\leq\mathcal{F}_{\mathrm{T}}
&\leq\mathcal{F}_{\mathrm{BT}}(\mathcal{P})\\
&\leq\mathcal{F}_{\mathrm{E\mbox{-}SGPR}}
\leq\log p(\mathbf{y}\mid\theta).
\end{aligned}
\]
The final inequality is the variational lower-bound property, with the
exact GP log marginal likelihood
\(
\log p(\mathbf{y}\mid\theta)
=\log\N(\mathbf{y};0,\Kff+\sigma^2\Id_N).
\)

In particular, SGPR and T-SGPR have the same training complexity,
\(O(M^3+NM^2)\). BT-SGPR additionally requires the block Cholesky
factorizations, giving \(O(M^3+NM^2+Nb^2)\); choosing \(b=O(M)\) retains the sparse training order
\(O(M^3+NM^2)\).

\paragraph{Stochastic optimization.}

Sparse variational methods are commonly expressed in uncollapsed form,
which also accommodates non-Gaussian likelihoods and allows stochastic
estimation of the bound \citep{hensman2013,hensman2015}. Rather than substituting the analytic optimum
of the inducing posterior, SVGP retains a free Gaussian distribution
\(
    q(\mathbf{u})=\N(\mathbf{m},\mathbf{S}),
\)
and trains the variational mean \(\mathbf{m}\), covariance \(\mathbf{S}\),
kernel hyperparameters, and usually the inducing points. Keeping the scaling
matrix explicit, the Gaussian-regression bound is
\citep{bui2025structured}
\[
\begin{aligned}
\mathcal{L}(q,\mathbf{M})
 &=-\KL[q(\mathbf{u})\Vert p(\mathbf{u})]\\
 &\quad+\sum_{i=1}^N\E_{q(\mathbf{u})}
 \log\N(y_i;\mathbf{k}_{i\mathbf{u}}\Kuu^{-1}\mathbf{u},\sigma^2)\\
 &\quad-\tfrac12\left\{
 \tr[(\Id_N+\sigma^{-2}\Dff)\mathbf{M}]
 -\log|\mathbf{M}|-N\right\},
\end{aligned}
\]
where \(\mathbf{k}_{i\mathbf{u}}\) is the \(i\)th row of \(\Kfu\).
The choices \(\mathbf{M}=\Id_N\), optimized diagonal \(\mathbf{M}\),
and optimized block-diagonal \(\mathbf{M}\) give SVGP, T-SVGP, and
BT-SVGP, respectively; we write \(\mathcal L_{\mathrm T}\) and
\(\mathcal L_{\mathrm{BT}}\) for the last two bounds.

Sampling one block \(\mathcal B_r\) uniformly and using it, scaled by \(B\),
for both the likelihood and the block log-determinant gives an unbiased
estimator of this bound \citep{bui2025structured};
Section~\ref{sec:svgp-experiments} separates the two samples for
geometrically selected blocks.

\paragraph{Prediction.}

For \(q(\mathbf{u})=\N(\mathbf{m},\mathbf{S})\), define
\(\mathbf{A}_\ast=\mathbf{k}_{\ast\mathbf{u}}\Kuu^{-1}\) and
\(d_{\ast\ast}=k_{\ast\ast}-\mathbf{k}_{\ast\mathbf{u}}\Kuu^{-1}\mathbf{k}_{\mathbf{u}\ast}\).
The standard SVGP predictor has latent mean and variance
\begin{equation}
    \mu_\ast=\mathbf{A}_\ast\mathbf{m},\qquad
    v_\ast^{\mathrm{SVGP}}
    =d_{\ast\ast}+\mathbf{A}_\ast\mathbf{S}\mathbf{A}_\ast^\top.
\label{eq:review-svgp-prediction}
\end{equation}
The prediction costs \(O(M)\) and
\(O(M^2)\) per test point, respectively, independently of \(N\).
All three models retain this form for the mean. Their variances have the
common representation \citep{bui2025tight,bui2025structured}
\begin{equation}
\begin{aligned}
    v_\ast
    &=v_\ast^{\mathrm{SVGP}}
      -\mathbf{d}_{\ast\mathbf{f}}\mathbf{V}_{\mathbf{ff}}
       \mathbf{d}_{\mathbf{f}\ast},\\
    \mathbf{V}_{\mathbf{ff}}
    &=\Dff^{-\top/2}(\Id-\mathbf{M})\Dff^{-1/2},
\end{aligned}
\label{eq:bui-predictive-variance}
\end{equation}
where \(v_\ast^{\mathrm{SVGP}}\) uses the \(q(\mathbf u)\) fitted by the
model under consideration and
\(\mathbf{d}_{\ast\mathbf{f}}=\mathbf{k}_{\ast\mathbf{f}}-\mathbf{k}_{\ast\mathbf{u}}\Kuu^{-1}\Kuf\).
For prediction we approximate the residual covariance by a block-diagonal
matrix, \(\Dff\approx\operatorname{blockdiag}_r\mathbf D_r\) with
\(\mathbf D_r=\Dff[\mathcal B_r,\mathcal B_r]\), using the final partition
for (R)BT-SVGP and (S)BT-SVGP and a partition from Algorithm~\ref{alg:sbt} for
T-SVGP. Substituting it into \(\mathbf V_{\mathbf{ff}}\) with the optimized
scaling \(\mathbf M\) gives
\(\mathbf V_{\mathbf{ff}}\approx\operatorname{blockdiag}_r\mathbf V_r\), where
\[
\mathbf V_r=
\begin{cases}
(\mathbf D_r+\sigma^2\Id)^{-1} & \text{BT-SVGP},\\
\mathbf D_r^{-1/2}\boldsymbol\Lambda_r\mathbf D_r^{-1/2} & \text{T-SVGP},
\end{cases}
\]
and \(\boldsymbol\Lambda_r\) is diagonal with entries
\([\Dff]_{ii}/([\Dff]_{ii}+\sigma^2)\), \(i\in\mathcal B_r\). For standard SVGP,
\(\mathbf M=\Id_N\) and the correction vanishes. Since
\(\mathbf V_{\mathbf{ff}}\) is now block diagonal, we place test inputs as in
PIC \citep{snelson2007}, the extension of PITC \citep{quinonero2005} to
test points: each test input is assigned, without its label, to the block
\(r\) whose representative is nearest in kernel distance, and only its
residual cross-covariance \(\mathbf d_{\ast r}\) with that block is kept,
\[
v_\ast=v_\ast^{\mathrm{SVGP}}-\mathbf d_{\ast r}\mathbf V_r\mathbf d_{r\ast}.
\]
The test input thus joins one block of a block-diagonal residual model, so
\(v_\ast\geq0\); for BT-SVGP the correction is the
variance reduction from conditioning the residual at \(x_\ast\) on the
noisy residuals of its block. After an \(O(N(M+b)^2)\) preparation of the
blocks, each prediction costs \(O(M^2+bM+b^2+N/b)\), against \(O(M^2)\) for
SVGP. Unlike PITC and PIC, where the block-diagonal residual defines an
approximate prior for both training and prediction and the PIC mean gains a
local term, the model here is the exact GP: the blocks enter training only
through the variational scaling, and the mean stays \(\mathbf A_\ast\mathbf m\).
We call the resulting T-SVGP, (R)BT-SVGP and (S)BT-SVGP predictors
\emph{block-diagonal predictors} (Appendix~\ref{app:hard-block-predictor}).
\section{Why the Partition Matters}

\begin{figure}[t]
    \centering
    \includegraphics[width=.85\columnwidth]{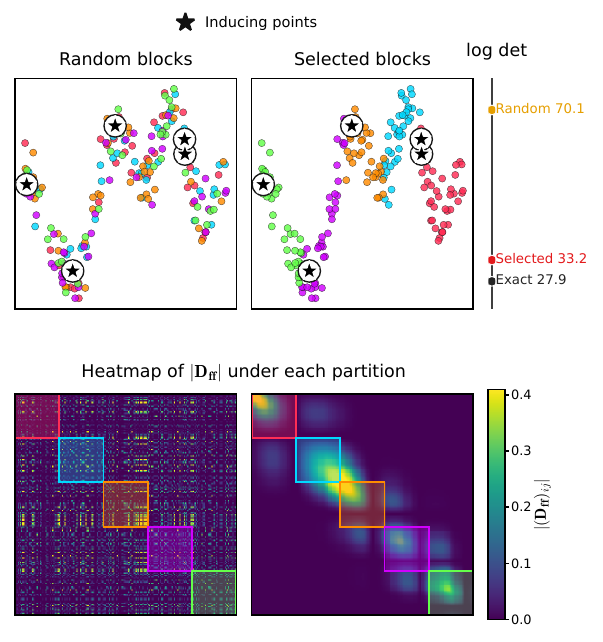}
    \caption{The same fitted Snelson residual under random blocks (left) and selected blocks (right), using five inducing points. Stars mark inducing points; colours denote block membership. Blocks have \(b=40\) points, and the log-det scale on the right compares the resulting penalties with the exact one.}

    \label{fig:snelson-residual-illustration}
\end{figure}
\label{sec:permutation}

The block penalty of BT-SGPR upper-bounds the dense log-determinant:
\[
\textstyle\sum_{r}\log|\Id+\sigma^{-2}\Dff[\mathcal{B}_r,\mathcal{B}_r]|
\geq\log|\Id_N+\sigma^{-2}\Dff|.
\]
The right-hand side is invariant under simultaneous row and column
permutations, whereas the left-hand side depends strongly on the partition
\(\mathcal{P}\). The partition is therefore a degree of freedom of the
bound, and we ask how to choose it so that the bound is as tight as possible.
Figure~\ref{fig:snelson-residual-illustration} illustrates this on the Snelson dataset \citep{snelson2005}: at the parameters of a fitted (S)BT-SGPR model, the same residual matrix is cut into random blocks and into the blocks selected by the algorithm of the next section. A good partition keeps most of the residual mass inside the diagonal blocks, which tightens the penalty.

For a partition \(\mathcal{P}\), write
\[
\begin{aligned}
    \mathbf{D}_{\mathcal{P}}
    &=
    \operatorname{blockdiag}_{r=1}^B
    \{\Dff[\mathcal{B}_r,\mathcal{B}_r]\},\\
    \mathbf{D}_{\mathcal{P}}^\perp
    &=
    \Dff-\mathbf{D}_{\mathcal{P}} .
\end{aligned}
\]
The block gap relative to the dense endpoint is
\[
\begin{aligned}
    \mathcal{E}(\mathcal{P})
    &=
    \sum_{r=1}^{B}
    \log\left|\Id_{|\mathcal{B}_r|}
    +\sigma^{-2}\Dff[\mathcal{B}_r,\mathcal{B}_r]\right|\\
    &\quad-\log|\Id_N+\sigma^{-2}\Dff|.
\end{aligned}
\]
\begin{proposition}[Uniform residual-gap bound]
\label{prop:leakage-uniform}
Let \(\Dff\succeq0\), \(\sigma^2>0\), and
\(\kappa=1+\sigma^{-2}\|\Dff\|_2\). Every partition satisfies
\begin{equation}
\frac{\|\mathbf D_{\mathcal P}^{\perp}\|_F^2}{2\sigma^4 \kappa^2}
\leq\mathcal E(\mathcal P)
\leq\frac{\|\mathbf D_{\mathcal P}^{\perp}\|_F^2}{2\sigma^4}.
\label{eq:uniform-gap}
\end{equation}
Consequently, at fixed continuous parameters,
\begin{equation}
\begin{aligned}
0&\leq\mathcal F_{\mathrm{E\mbox{-}SGPR}}-\mathcal F_{\mathrm{BT}}(\mathcal P)\\
 &=\tfrac12\mathcal E(\mathcal P)
 \leq\frac{\|\mathbf D_{\mathcal P}^{\perp}\|_F^2}{4\sigma^4}.
\end{aligned}
\label{eq:elbo-gap}
\end{equation}
\end{proposition}
The proof in Appendix~\ref{app:gap-proof} interpolates between
\(\mathbf D_{\mathcal P}\) and \(\Dff\). The first-order
change vanishes because the added interactions are off-block; bounded
second derivatives give the quadratic scale.

For \(N\geq2\), random partitions are weak for this target. If \(\mathcal{P}_{\mathrm R}\) is a uniform
equal-size random partition with block size \(b\), then any pair is separated
with probability \((N-b)/(N-1)\) (Proposition~\ref{prop:random-offblock},
Appendix~\ref{app:random}), so
\[
    \E \|\mathbf{D}_{\mathcal{P}_{\mathrm R}}^\perp\|_F^2
    =
    \frac{N-b}{N-1}
    \sum_{i\neq j}[\Dff]_{ij}^2.
\]
When \(b\ll N\), almost all off-diagonal residual mass is outside random
blocks. This motivates nonrandom selection when residual interactions
are sufficiently structured.

\section{Selected Blocks}
\label{sec:selection}
Section~\ref{sec:permutation} turns the choice of partition into a
clustering problem: keep as much residual mass \([\Dff]_{ij}^2\) as possible
inside the blocks. Solving it directly requires the dense residual. We use
instead the geometry of the kernel, through the kernel distance
\citep{rasmussen2006}
\begin{equation}
d_\theta^2(x,z)=k_\theta(x,x)+k_\theta(z,z)-2k_\theta(x,z),
\label{eq:anchor-distance}
\end{equation}
and the inducing points as \emph{anchors}.

\paragraph{Why inducing points are good anchors.}
Let \(\phi\) be a feature map of \(k_\theta\), \(P_Z\) the orthogonal
projector onto \(\operatorname{span}\{\phi(z_1),\ldots,\phi(z_M)\}\), and
\(r_Z(x)=(I-P_Z)\phi(x)\) the unexplained feature. Exact inducing
conditioning gives the residual kernel \(D(x,y)=\langle r_Z(x),r_Z(y)\rangle\),
with \([\Dff]_{ij}=D(x_i,x_j)\), and \(r_Z(z_m)=0\). Since an orthogonal projection is contractive,
\[
\begin{aligned}
  \|r_Z(x)-r_Z(y)\|&\leq d_\theta(x,y),\\
  D(x,x)&\leq d_\theta^2(x,z_m),\\
  2D(x,y)&\geq D(x,x)+D(y,y)-d_\theta^2(x,y).
\end{aligned}
\]
Inducing points are therefore zeros of the residual, assigning a point to
its nearest inducing point minimizes an upper bound on its residual
variance, and two close points with appreciable residual variances have a
strong positive residual interaction. The kernel distance is thus a
cheap proxy for the residual geometry of the block
objective; Theorem~\ref{thm:geometric-recovery} gives conditions under
which it yields the optimal partition.

\paragraph{Algorithm.}
Algorithm~\ref{alg:sbt} builds the partition in four steps. (i) Each point
joins the cell of its nearest anchor (lines 1--2). Since \(k_\theta(x_i,x_i)\) does not
depend on the anchor,
\[
a(i)=\arg\min_{m}\{[\Kuu]_{mm}-2[\Kfu]_{im}\},
\]
which reuses quantities already computed in training and costs \(O(NM)\),
without forming the residual. (ii) The anchors are ordered along a
nearest-neighbour chain (line 3), so that neighbouring cells become consecutive.
(iii) Within each cell, a boundary score orders the points from the side of
the previous anchor to the side of the next one, so that a group lying
between two anchors, which the cells split in two, stays contiguous
(lines 4--5). (iv) The sequence is cut into blocks of size \(b\) (line 6), which stay balanced even
when cells have different sizes. The search requires
\(O(NM+M^2+N\log N)\) kernel-distance work per refresh. The partition and
ordering require \(O(N)\) storage, excluding kernel/distance caches and
temporary workspace; materializing all point--anchor and anchor--anchor
quantities, as our implementation does, requires an additional
\(O(NM+M^2)\) storage.

\providecommand{\algcost}[1]{\unskip\nobreak\hfill\penalty50\hskip1em\null\nobreak\hfill\ensuremath{#1}}

\begin{algorithm}[t]
\caption{Inducing-Anchored Selected-Block Tightening}
\label{alg:sbt}
\begin{algorithmic}[1]
\REQUIRE Training inputs \(\mathbf{X}=\{x_i\}_{i=1}^N\), inducing points
\(\mathbf{Z}=\{z_m\}_{m=1}^M\), hyperparameters \(\theta\), block size \(b\).
\ENSURE Partition \(\mathcal{P}_{\mathrm{(S)BT}}=\{\mathcal{B}_r\}_{r=1}^{B}\)
of \(\{1,\dots,N\}\), with \(B=\lceil N/b\rceil\).
\STATE \textbf{Distances.} Reusing \(\Kfu\), \(\Kuu\) and the diagonals of
\(\Kff\) and \(\Kuu\) cached during training,
compute \(d_\theta^2(x_i,z_m)\) for every point and anchor, and
\(d_\theta^2(z_m,z_n)\) for every pair of anchors
(Eq.~\eqref{eq:anchor-distance}).
\algcost{O(NM+M^2)}
\STATE \textbf{Cells.} Assign each point to its nearest anchor,
\(a(i)=\arg\min_m d_\theta(x_i,z_m)\). The points with \(a(i)=m\) form the
cell of \(z_m\).
\algcost{O(NM)}
\STATE \textbf{Chain.} Arrange the anchors along a path: start from
\(z_1\) and repeatedly move to the nearest anchor in \(d_\theta\) not yet
visited. This gives the order \(z_{c_1},\dots,z_{c_M}\), in which
consecutive anchors are close. Let \(z_{m^-}\) and \(z_{m^+}\) be the anchors
just before and just after \(z_m\) in this order.
\algcost{O(M^2)}
\STATE \textbf{Boundary score.} For each point, with \(m=a(i)\), let
\(s_i=d_\theta^2(x_i,z_{m^-})-d_\theta^2(x_i,z_{m^+})\). It is negative when
\(x_i\) is closer to the previous anchor than to the next one. In the first
cell of the chain use \(s_i=-d_\theta^2(x_i,z_{m^+})\), in the last
\(s_i=d_\theta^2(x_i,z_{m^-})\), and if \(M=1\), \(s_i=d_\theta^2(x_i,z_1)\).
\algcost{O(N)}
\STATE \textbf{Ordering.} List the cells in chain order and, within each
cell, its points by increasing \(s_i\): from the side facing the previous
cell to the side facing the next one. Break ties by
\(d_\theta(x_i,z_{a(i)})\), and call the resulting sequence
\(\pi=(\pi_1,\dots,\pi_N)\).
\algcost{O(N\log N)}
\STATE \textbf{Blocks.} Cut \(\pi\) into consecutive pieces of \(b\) points,
\(\mathcal{B}_r=\{\pi_j:\lceil j/b\rceil=r\}\).
\algcost{O(N)}
\RETURN \(\mathcal{P}_{\mathrm{(S)BT}}\).
\end{algorithmic}
\end{algorithm}

\paragraph{Refreshes.}
Selecting the blocks does not create a new bound: at fixed \(\theta\) and
\(\mathbf Z\), (S)BT-SGPR and (S)BT-SVGP are the BT-SGPR and BT-SVGP bounds
evaluated on the partition \(\mathcal{P}_{\mathrm{(S)BT}}(\theta,\mathbf Z)\). Since \(\theta\) and \(\mathbf Z\) change
during training, most strongly early on, we refresh the partition
periodically. At each refresh, Algorithm~\ref{alg:sbt} recomputes the
partition at the current \(\theta\) and \(\mathbf Z\), and training resumes from
these parameters with the updated bound. Proposition~\ref{prop:partition-exchange}
(Appendix~\ref{app:exchange}) bounds the resulting change in the ELBO and
gives a condition under which the refresh tightens the bound.

\begin{figure}[!h]
\centering
\includegraphics[width=.9\columnwidth]{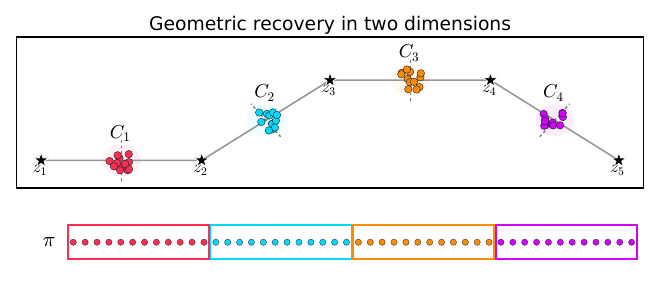}
\caption{Two-dimensional RBF example of Theorem~\ref{thm:geometric-recovery}.
The ordering \(\pi\) (bottom) returns each coloured group as one block;
stars are inducing points, dashed lines cell boundaries.}
\label{fig:geometric-recovery-main}
\end{figure}

\paragraph{The ideal case.}
Suppose the data form compact groups, each lying between two consecutive
inducing points. The cells split every group in two, but the chain and the
boundary score put the two halves next to each other before the cut
(Figure~\ref{fig:geometric-recovery-main}). The groups are not given to the
algorithm; Theorem~\ref{thm:geometric-recovery} shows that it recovers them
and, under a further condition, that they form the best balanced partition.

\begin{theorem}[Geometric recovery]
\label{thm:geometric-recovery}
At fixed \(\theta\) and \(\mathbf Z\), assume \(M\geq2\), \(\sigma^2>0\),
\(\Kuu\succ0\), and \(k_\theta(x_i,x_i)>0\) for every observation. Label the inducing points so that the chain
of Algorithm~\ref{alg:sbt} starts at \(z_1\). For each \(m<M\), let \(C_m\)
contain exactly the observations \(x\) satisfying
\[
\max_{a\in\{m,m+1\}}d_\theta(x,z_a)<d_\theta(x,z_j),
\]
for every \(j\notin\{m,m+1\}\). Suppose these groups cover all observations
and \(|C_m|=b\geq2\), so \(N=(M-1)b\). They are disjoint, although a
point may be equally close to its two adjacent anchors and assigned to
either. Assume that
\begin{enumerate}
\item[(i)] \(d_\theta(z_m,z_{m+1})<d_\theta(z_m,z_j)\) for every
\(j>m+1\);
\item[(ii)] \(d_\theta(x,y)\leq h\) for observations \(x,y\) in the same
group \(C_m\). For \(\delta\geq0\),
\(|k_\theta(x,y)|\leq\delta\sqrt{k_\theta(x,x)k_\theta(y,y)}\)
for observations in different groups, for distinct inducing points, and
for every observation--inducing pair, including coincident input locations.
\end{enumerate}
Then Algorithm~\ref{alg:sbt} returns \(\mathcal Q=\{C_1,\ldots,C_{M-1}\}\).
If moreover \(h^2<2\min_ik_\theta(x_i,x_i)\), there is \(\delta_0>0\) such
that, for \(\delta<\delta_0\), \(\mathcal Q\) has a strictly larger ELBO,
collapsed or not, than every other partition into blocks of size \(b\).
\end{theorem}
Trained inducing points need not produce this configuration, but the case
shows what the algorithm is built to find. Since the inducing points are zeros
of the residual \(D\), the blocks act only on the structure that the inducing
points leave unexplained, and the theorem shows that the algorithm finds such
local groups and keeps each one in a single block.
Appendix~\ref{app:recovery} proves the theorem, gives an explicit family in
which both \(b/N\) and \(M/N\) vanish, and shows when the hypotheses hold for
the kernels of our experiments.

\section{Experiments}
\label{sec:experiments}

We ask three questions: does the selected partition tighten the bound and
improve predictions over random blocks of the same size; how close does it
come to a dense graph partitioner; and do its gains survive when the
structured predictor is replaced by a standard sparse one?

\subsection{Collapsed SGPR}
\label{sec:sgpr}

\begin{figure}[H]
    \centering
    \captionsetup{width=\columnwidth}
    \includegraphics[width=.85\columnwidth]{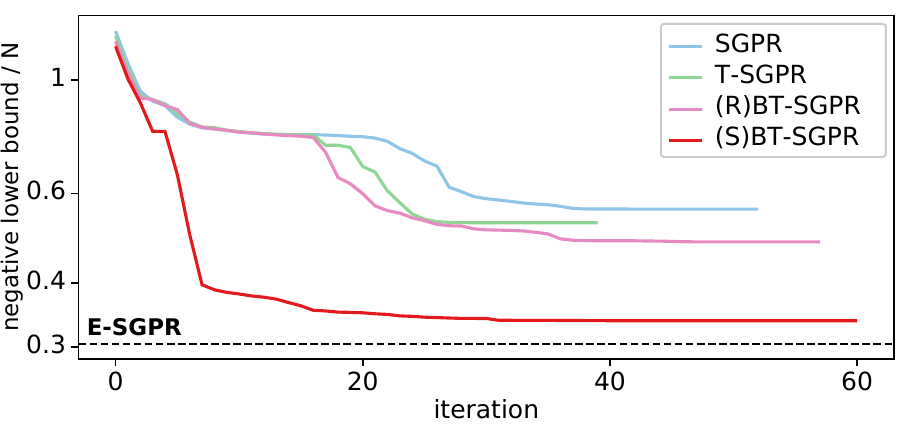}
    \par\medskip
    \includegraphics[width=.85\columnwidth]{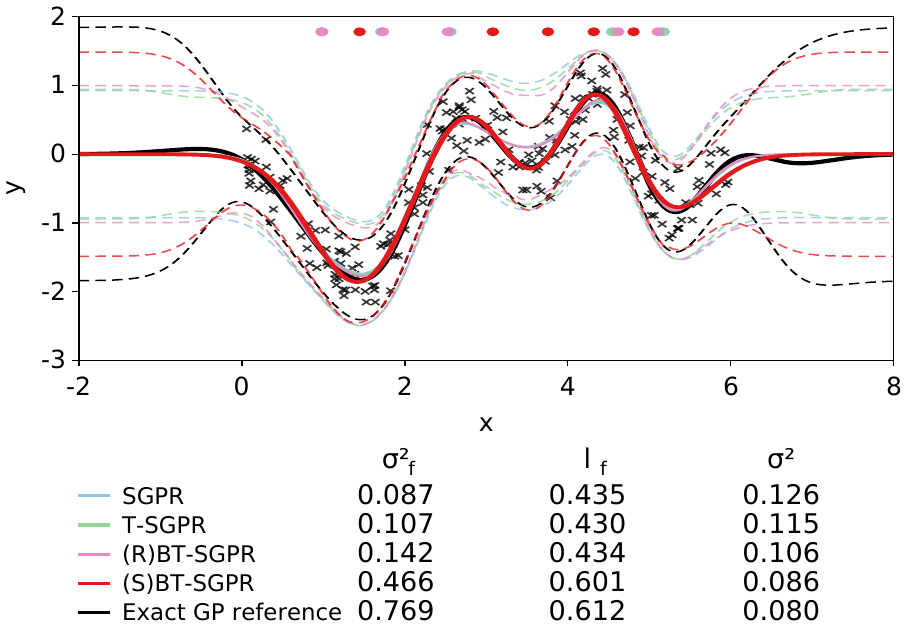}
    \caption{Snelson: training bounds (top; dashed: E-SGPR); predictive
    fits with the block-diagonal predictor for T-SGPR, (R)BT-SGPR and (S)BT-SGPR and fitted hyperparameters
    (bottom).}
    \label{fig:snelson}
\end{figure}

We first fit SGPR, T-SGPR, (R)BT-SGPR and (S)BT-SGPR to the Snelson dataset
with \(M=5\) inducing points (Figure~\ref{fig:snelson}). The tighter the
bound, the closer the fitted hyperparameters are to those of the exact GP,
and (S)BT-SGPR comes closest, in line with
\citet{bui2025tight,bui2025structured}.

\begin{table}[!ht]
\caption{RBF--Kin40k: \(M=512\), \(N=5000\), 2000 test points and \(b=500\). Test RMSE/NLL use the block-diagonal predictor of Eq.~\eqref{eq:hard-block-correction}; mean (SE in parentheses) over ten seeds. Search is the partition-build time in ms. Bold: best non-dense method; \(^\dagger\)dense reference.}
\label{tab:kin40k-sgpr}
\centering\footnotesize
\setlength{\tabcolsep}{3pt}
\begin{tabular}{@{}lrrr@{}}
\toprule
Method & RMSE & NLL & Search (ms) \\
\midrule
SGPR & $0.229(2)$ & $-0.041(7)$ & --- \\
T-SGPR & $0.205(2)$ & $-0.146(7)$ & --- \\
(R)BT-SGPR & $0.197(2)$ & $-0.179(7)$ & --- \\
(S)BT-SGPR & \textbf{$0.185(2)$} & \textbf{$-0.240(7)$} & 6.07 \\
\midrule
(METIS)BT-SGPR$^\dagger$ & $0.179(2)$ & $-0.296(6)$ & 1034.92 \\
Exact GP$^\dagger$ & $0.151(2)$ & $-0.618(14)$ & --- \\
\bottomrule
\end{tabular}
\end{table}

We then fit the same methods on Kin40k with an RBF kernel (\(N=5000\),
\(M=512\), \(b=500\), L-BFGS-B). The test metrics follow the hierarchy of the
bounds (Table~\ref{tab:kin40k-sgpr}). At the same block size, selecting the
partition instead of drawing it at random lowers the RMSE from 0.197 to 0.185
and the NLL from \(-0.179\) to \(-0.240\). As a reference we also partition
the dense residual graph with METIS \citep{karypis1998}, with edge weights
\([\Dff]_{ij}^{2}\), so that its cut directly targets
\(\|\mathbf{D}_{\mathcal P}^{\perp}\|_{F}^{2}\). The resulting (METIS)BT-SGPR does slightly better
(0.179 and \(-0.296\)), but its search takes 1035\,ms per build, against
6\,ms for Algorithm~\ref{alg:sbt}.

\subsection{Sparse predictor transplant and refinement}
\label{sec:sgpr-transplant}
Dropping the structured correction at test time,
\(q(f\mid\mathbf{u})\approx p(f\mid\mathbf{u})\), transfers the
hyperparameters and inducing points of a structured model (the \emph{donor}) to a standard
SGPR predictor (a \emph{transplant}). We evaluate the transplanted model as it is
(\emph{direct sparse}) and after optimizing \(\mathbf Z\) on the SGPR ELBO until
convergence, with the kernel and noise hyperparameters fixed
(\emph{refined sparse}). Refinement slightly worsens the T-SGPR and
(R)BT-SGPR donors but improves the (S)BT-SGPR donor, whose NLL goes from
\(-0.158\) to \(-0.164\) (its RMSE rises from 0.185 to 0.190); (S)BT-SGPR is the best sparse predictor in both
variants (Table~\ref{tab:kin40k-sgpr-refined},
Appendix~\ref{app:experiments}). This suggests that selected-block training
learns hyperparameters that transfer well to a standard SGPR predictor.

\subsection{Stochastic SVGP}
\label{sec:svgp-experiments}

\paragraph{Setup.}
We train SVGP, T-SVGP, (R)BT-SVGP and (S)BT-SVGP with four kernels (RBF,
Mat\'ern-5/2, Amplitude-RBF and Gibbs) on eight datasets, with
\(M\in\{64,128,256,512\}\) and ten seeds per configuration. Within a kernel,
all methods share the same epoch budget: 350 for RBF and Mat\'ern-5/2, 500
for Amplitude-RBF, and 300 after an RBF warm start for Gibbs. (R)BT-SVGP and
(S)BT-SVGP are trained with the split estimator described at the end of this
section.

\paragraph{Accuracy.}
Figure~\ref{fig:svgp-final} shows RBF and Amplitude-RBF on Pol
\citep{weiss1995,vanschoren2014openml}, Superconductivity
\citep{hamidieh2018,uciSuperconductivity2018}, Kin40k \citep{schwaighofer2003}
and Protein \citep{uciProtein2013}; the other kernels and datasets are in
Appendix~\ref{app:svgp-frozen}, the protocol in
Appendix~\ref{app:experiments}. With the block-diagonal predictors (solid lines),
(S)BT-SVGP has the lowest RMSE and the lowest NLL in all 32 displayed
configurations, improving on (R)BT-SVGP by 0.006 to 0.020 in RMSE and by
0.015 to 0.117 in NLL.

\paragraph{Cost.}
(R)BT-SVGP and (S)BT-SVGP have the same block-factorization cost, and selection adds
little: on RBF--Kin40k at \(M=512\), building the selected partitions takes
1.74\,s in total against 114.85\,s of optimization (1.52\%, means over ten
seeds), while (R)BT-SVGP optimization takes 114.99\,s. The price is paid at
prediction time (Section~\ref{sec:bounds}).

\paragraph{Sparse transplant.}
We therefore repeat the transplant of Section~\ref{sec:sgpr-transplant}:
each donor's hyperparameters are fixed, and the inducing points and
\(q(\mathbf{u})\) are refined under the standard SVGP objective for 300
epochs (dashed lines). The (S)BT-SVGP donors still give the lowest RMSE in all
32 configurations and the lowest NLL in 19. The 13 exceptions are all at
\(M\leq256\), ten of them on Kin40k and Pol, and in each of them the best
sparse NLL comes from the (R)BT-SVGP donor; at \(M=512\) the (S)BT-SVGP donor is best
everywhere. This is consistent with structured training learning more
useful hyperparameters; at small budgets, compressing the donated model
into the sparse predictor can lose part of the advantage.

\clearpage\onecolumn
\begin{landscape}
\begin{figure}[p]
\centering
\includegraphics[width=.99\linewidth,height=.83\textheight,keepaspectratio]{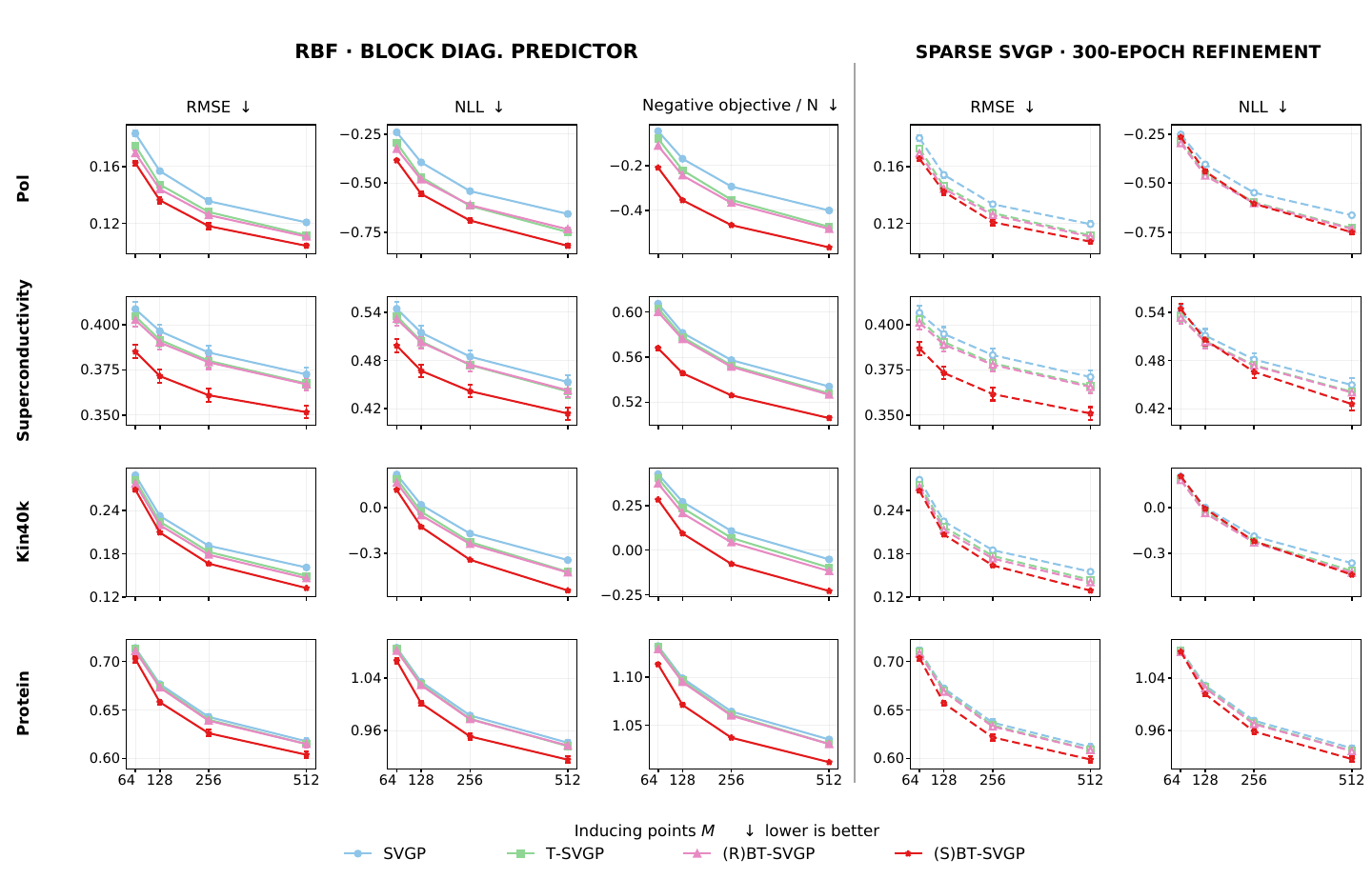}
\caption{RBF on Pol, Superconductivity, Kin40k and Protein. Left three columns:
RMSE, NLL and negative final objective per datum of the block-diagonal
predictors (Section~\ref{sec:bounds}; solid); right two columns: RMSE and NLL of
the sparse SVGP after 300-epoch transplant refinement (dashed). Means and one
standard error over ten seeds; lower is better. The objectives are
method-specific stochastic estimates, not a common marginal likelihood.
Amplitude-RBF follows on the next page; the full campaign is in
Appendix~\ref{app:svgp-frozen}.}
\label{fig:svgp-final}
\end{figure}
\begin{figure}[p]
\ContinuedFloat
\centering
\includegraphics[width=.99\linewidth,height=.83\textheight,keepaspectratio]{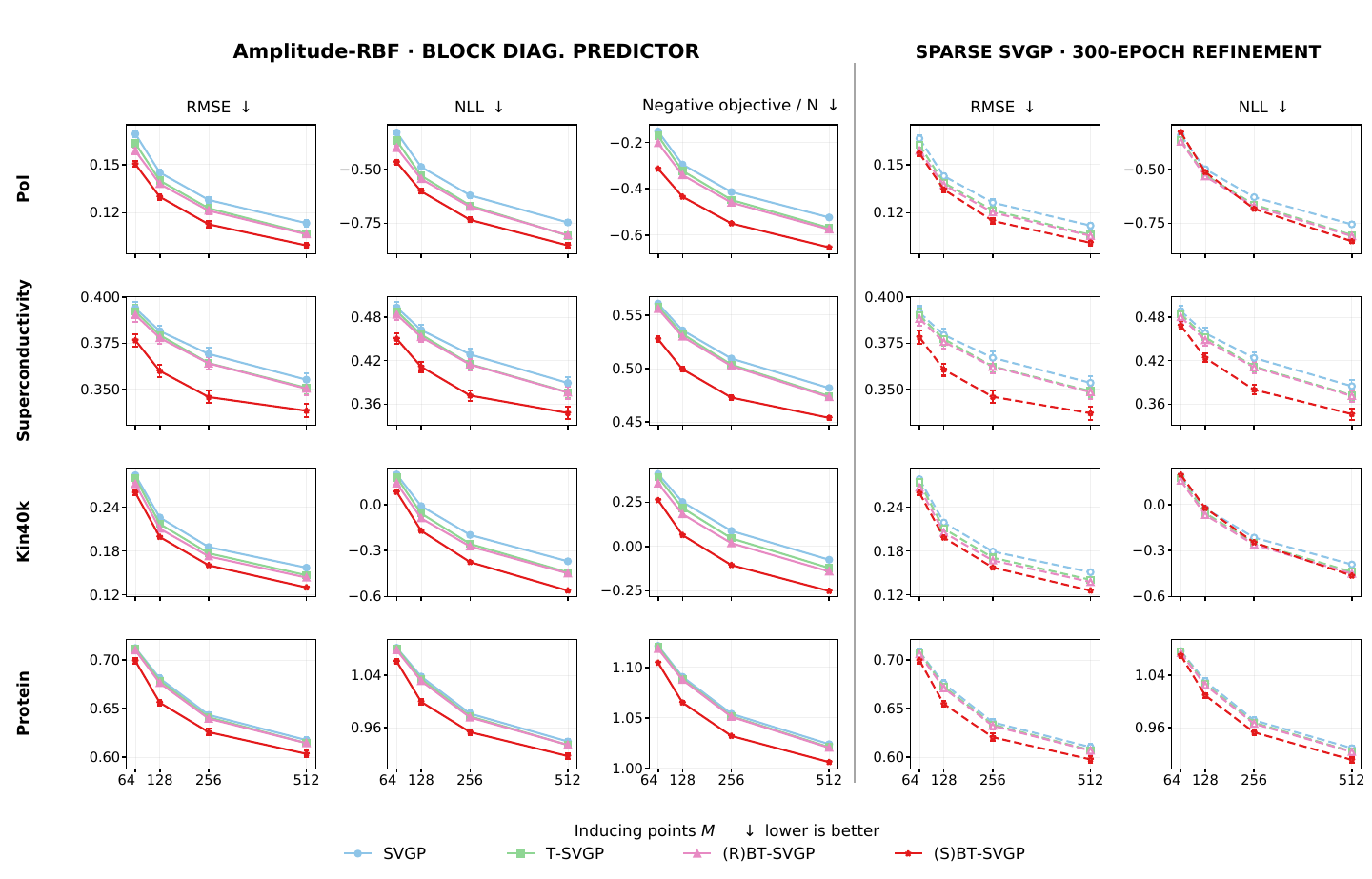}
\caption{(Continued.) Amplitude-RBF on Pol, Superconductivity, Kin40k and
Protein, with the same layout.}
\end{figure}
\end{landscape}
\twocolumn
\begin{figure}[H]
    \centering
    \includegraphics[width=.75\columnwidth]{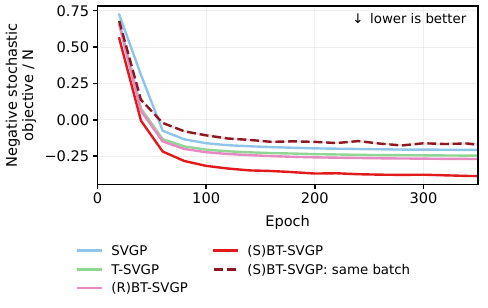}
    \caption{Negative stochastic objective during training on RBF--Pol,
    \(M=128\), ten seeds. Separate likelihood and block batches converge
    better than a single shared batch for (S)BT-SVGP.}
    \label{fig:svgp-batch-ablation}
\end{figure}

\paragraph{Split estimator.}
The BT-SVGP bound is the T-SVGP bound plus one gain per block,
\[
\begin{aligned}
\mathcal L_{\mathrm{BT}}&=\mathcal L_{\mathrm T}+\tfrac12\textstyle\sum_{r=1}^{B}g_r,\\
g_r&=\textstyle\sum_{i\in\mathcal B_r}\log\bigl(1+\sigma^{-2}[\Dff]_{ii}\bigr)\\
&\quad-\log\bigl|\Id+\sigma^{-2}\Dff[\mathcal B_r,\mathcal B_r]\bigr|.
\end{aligned}
\]
By Hadamard's inequality \(g_r\geq0\): it measures how much block
\(\mathcal B_r\) tightens the T-SVGP diagonal penalty by modelling the
residual correlations among its points. Selected blocks are localized in input space, so sampling one block for
both terms (Section~\ref{sec:bounds}) would make every update see a single
region of the data. We sample the two terms separately,
\[
\widehat{\mathcal L}_{\mathrm{split}}
=\widehat{\mathcal L}_{\mathrm T}(\mathcal B)+\frac{B}{2K}\sum_{k=1}^{K}g_{r_k},
\]
where \(\widehat{\mathcal L}_{\mathrm T}(\mathcal B)\) is the minibatch
estimate of \(\mathcal L_{\mathrm T}\) on a random likelihood batch
\(\mathcal B\) and \(r_1,\ldots,r_K\) are \(K\) random blocks; under uniform
sampling this is unbiased at fixed parameters, and in practice batches and
blocks are traversed in separate, uniformly weighted cycles.
Separating the two samples is not the same as doubling the batch: the
likelihood batch needs only the diagonal of \(\Dff\), and the
\(b\times b\) Cholesky factors are computed for the \(K\) blocks alone, so
the split adds \(O(|\mathcal B|M^2)\) to a shared batch, a factor close to
one, whereas doubling a shared batch doubles the Cholesky work
(Appendix~\ref{app:svgp}). We use \(|\mathcal B|=1000\) and \(K=2\) blocks of size
\(b=500\). (R)BT-SVGP and (S)BT-SVGP differ only in their blocks; SVGP and T-SVGP
use the likelihood batch alone. Separate batches converge better than a
shared one (Figure~\ref{fig:svgp-batch-ablation}).

\section{Conclusion}
\label{sec:conclusion}

Block-structured sparse GP bounds contain an internal degree of freedom:
the partition used in the residual log-determinant approximation.
Our results show that choosing this partition can substantially improve
performance. Building on existing block-structured bounds, we develop a
scalable block-search procedure and a doubly stochastic implementation
that stabilizes training. Finally, we use these models as a
parameter-learning stage for subsequent sparse SVGP refinement.
The resulting comparisons show improvements in RMSE and NLL across
datasets and kernels.

\paragraph{Limitations.}

Our guarantees are sufficient conditions at a fixed state; outside them,
selected blocks need not beat random ones, and tighter bounds need not
improve every test metric. The block-diagonal predictor remains
costly, and sparsification trades its covariance correction for cheaper
prediction. Hyperparameters learned under the block-structured objective need not be
optimal for the sparse predictor. Removing the covariance correction can
worsen NLL (Gibbs on Electricity,
Figure~\ref{fig:campaign-gibbs-additional}; Pol and Kin40k at small \(M\));
our results do not establish a reduction in statistical estimation bias. The objectives are non-convex, so numerical stopping criteria
do not certify global optimality.

\subsection*{AI Use Statement}
Generative AI tools were used to assist with developing and discussing
theoretical arguments and proofs; designing and implementing experiments,
training and evaluation code, and diagnostic checks; interpreting results;
translating and editing text; preparing figures and tables; and identifying
relevant literature. This assistance included work on the block-selection
motivation and the experimental workflows. The authors remain responsible
for the correctness, originality, attribution, and reproducibility of all
claims, proofs, code, and reported results.
The tools used were OpenAI Codex and Anthropic Claude. Generative AI was not
used to generate synthetic data, to clean datasets, or for qualitative
analysis. The authors verified these outputs as follows: they checked every
proof by hand; they re-ran the experiments with the released code and compared
every reported number with the stored result files; they checked every
reference against the original publication; and they read and revised all text.

\bibliography{references}

\clearpage
\appendix
\onecolumn
\thispagestyle{empty}
\aistatstitle{Geometry-Aware Block Selection for Sparse Variational Gaussian Processes:\\Supplementary Materials}
\raggedbottom
\section{Proofs}
\label{app:proofs}
Throughout, \(\mathbf D=\Dff\succeq0\), \(\mathbf D_{\mathcal P}\) keeps the diagonal blocks
of a partition \(\mathcal P\), \(\mathbf D_{\mathcal P}^\perp=\mathbf D-\mathbf D_{\mathcal P}\)
as in Section~\ref{sec:permutation}, and
\(J(\mathcal P)=\log|\Id+\sigma^{-2}\mathbf D_{\mathcal P}|\). At fixed
\((\theta,\mathbf Z,q(\mathbf u))\), the partition enters both the BT-SGPR bound
\(\mathcal F_{\mathrm{BT}}(\mathcal P)\) and the
BT-SVGP bound \(\mathcal L_{\mathrm{BT}}\) only through \(-\tfrac12J(\mathcal P)\); every
comparison below therefore holds for both, and we write
\(\mathcal L_{\mathrm{BT}}(\mathcal P)\).

\subsection{Proof of Proposition~\ref{prop:leakage-uniform}}
\label{app:gap-proof}
Let \(\mathbf R=\mathbf D_{\mathcal P}^\perp\), \(\mathbf A_t=\Id+\sigma^{-2}(\mathbf D_{\mathcal P}+t\mathbf R)\) and
\(f(t)=-\log|\mathbf A_t|\) for \(t\in[0,1]\), so that
\(\mathcal E(\mathcal P)=f(1)-f(0)\). Each block of \(\mathbf D_{\mathcal P}\) is a
principal submatrix of \(\mathbf D\), hence \(\|\mathbf D_{\mathcal P}\|_2\leq\|\mathbf D\|_2\) and
\(\Id\preceq \mathbf A_t\preceq \kappa\Id\). Differentiating,
\[
f'(t)=-\sigma^{-2}\tr(\mathbf A_t^{-1}\mathbf R),\qquad
f''(t)=\sigma^{-4}\tr(\mathbf A_t^{-1}\mathbf R\mathbf A_t^{-1}\mathbf R)=\sigma^{-4}\|\mathbf A_t^{-1/2}\mathbf R\mathbf A_t^{-1/2}\|_F^2.
\]
\(\mathbf A_0^{-1}\) is block diagonal and \(\mathbf R\) vanishes on the diagonal blocks, so
\(f'(0)=0\). From \(\Id\preceq \mathbf A_t\preceq \kappa\Id\),
\(\|\mathbf R\|_F/\kappa\leq\|\mathbf A_t^{-1/2}\mathbf R\mathbf A_t^{-1/2}\|_F\leq\|\mathbf R\|_F\). Taylor's
formula with integral remainder,
\(\mathcal E(\mathcal P)=\int_0^1(1-t)f''(t)\,dt\), with
\(\int_0^1(1-t)\,dt=\tfrac12\), gives Eq.~\eqref{eq:uniform-gap}, and
Eq.~\eqref{eq:elbo-gap} follows because the ELBO gap is
\(\tfrac12\mathcal E(\mathcal P)\). Since \(\tr(\mathbf A_0^{-1}\mathbf A_1)=N\), the gap is
also \(\mathcal E(\mathcal P)=2\KL[\N(0,\mathbf A_1)\Vert\N(0,\mathbf A_0)]\): the dependence
lost by replacing the noisy residual with the product of its block
marginals.\hfill\(\square\)

\subsection{Random Partitions}
\label{app:random}
\begin{proposition}
\label{prop:random-offblock}
For \(N\geq2\), cut a uniform random permutation of \(\{1,\ldots,N\}\) into blocks of sizes
\(b_1,\ldots,b_B\). Then
\[
\E\|\mathbf D_{\mathcal P}^\perp\|_F^2=p\sum_{i\neq j}[\Dff]_{ij}^2,\qquad
p=1-\frac{\sum_rb_r(b_r-1)}{N(N-1)},
\]
and \(p=(N-b)/(N-1)\) when every block has size \(b\) (so \(b\mid N\)).
\end{proposition}
\noindent\textit{Proof.}
An ordered pair \(i\neq j\) occupies two uniformly random distinct positions,
and \(\sum_rb_r(b_r-1)\) of the \(N(N-1)\) ordered position pairs lie in a
common block.\hfill\(\square\)

\subsection{Partition-Exchange Bound}
\label{app:exchange}
For an incumbent partition \(\mathcal P\) and a new partition \(\mathcal Q\),
let \(\mathcal I_+\) (\(\mathcal I_-\)) be the ordered pairs \(i\neq j\) that
share a block of \(\mathcal Q\) but not of \(\mathcal P\) (of \(\mathcal P\)
but not of \(\mathcal Q\)), and
\[
G_\pm=\sum_{(i,j)\in\mathcal I_\pm}[\Dff]_{ij}^2,\qquad
\kappa_{\mathcal P}=1+\sigma^{-2}\|\mathbf D_{\mathcal P}\|_2,\qquad
\kappa_{\mathcal Q}=1+\sigma^{-2}\|\mathbf D_{\mathcal Q}\|_2.
\]
\(G_+\) and \(G_-\) are the residual interactions that the new partition
gains and loses.

\begin{proposition}[Partition-exchange bound]
\label{prop:partition-exchange}
For \(\sigma^2>0\),
\begin{equation}
\frac{G_+/\kappa_{\mathcal Q}^2-G_-}{4\sigma^4}
\leq\mathcal L_{\mathrm{BT}}(\mathcal Q)-\mathcal L_{\mathrm{BT}}(\mathcal P)
\leq\frac{G_+-G_-/\kappa_{\mathcal P}^2}{4\sigma^4}.
\label{eq:exchange-bound}
\end{equation}
In particular, a refresh with \(G_+\geq\kappa_{\mathcal Q}^2G_-\) does not
decrease the ELBO, and strictly increases it under strict inequality.
\end{proposition}
\noindent\textit{Proof.}
\label{app:exchange-proof}
Let \(\mathcal H=\mathcal P\wedge\mathcal Q\) be the common refinement.
Keeping the \(\mathcal P\)-blocks of \(\mathbf D_{\mathcal Q}\) gives
\(\mathbf D_{\mathcal H}\), and so does keeping the \(\mathcal Q\)-blocks of
\(\mathbf D_{\mathcal P}\); hence \(\|\mathbf D_{\mathcal Q}-\mathbf D_{\mathcal H}\|_F^2=G_+\) and
\(\|\mathbf D_{\mathcal P}-\mathbf D_{\mathcal H}\|_F^2=G_-\). Proposition~\ref{prop:leakage-uniform},
applied to \(\mathbf D_{\mathcal Q}\succeq0\) with partition \(\mathcal P\) and to
\(\mathbf D_{\mathcal P}\succeq0\) with partition \(\mathcal Q\), gives
\[
\frac{G_+}{2\sigma^4\kappa_{\mathcal Q}^2}\leq J(\mathcal H)-J(\mathcal Q)\leq\frac{G_+}{2\sigma^4},
\qquad
\frac{G_-}{2\sigma^4\kappa_{\mathcal P}^2}\leq J(\mathcal H)-J(\mathcal P)\leq\frac{G_-}{2\sigma^4}.
\]
The claim follows from
\(\mathcal L_{\mathrm{BT}}(\mathcal Q)-\mathcal L_{\mathrm{BT}}(\mathcal P)
=\tfrac12[(J(\mathcal H)-J(\mathcal Q))-(J(\mathcal H)-J(\mathcal P))]\).\hfill\(\square\)

\subsection{Residual Separation}
\label{app:ascent-proof}
\begin{lemma}[Residual separation]
\label{lem:separation}
Let \(\mathcal P\neq\mathcal Q\) be partitions with the same block sizes, the
largest being \(b\). Let \(u,v\geq0\), and suppose that \(|[\Dff]_{ij}|\geq u\) for \(i\neq j\) in a
common block of \(\mathcal Q\) and \(|[\Dff]_{ij}|\leq v\) otherwise. Then
\(m=|\mathcal I_+|>0\) and
\begin{equation}
\mathcal L_{\mathrm{BT}}(\mathcal Q)-\mathcal L_{\mathrm{BT}}(\mathcal P)
\geq\frac{m}{4\sigma^4}\Bigl(\frac{u^2}{\kappa_{\mathcal Q}^2}-v^2\Bigr),
\qquad
\kappa_{\mathcal Q}\leq1+\frac{b\max_i[\Dff]_{ii}}{\sigma^2}.
\label{eq:separation-improvement}
\end{equation}
Hence, if \(u>\kappa_{\mathcal Q}v\), \(\mathcal Q\) has a strictly larger
ELBO than every other partition with its block sizes.
\end{lemma}
\noindent\textit{Proof.}
Both partitions contain \(\sum_rb_r(b_r-1)\) ordered within-block pairs, so
\(|\mathcal I_+|=|\mathcal I_-|=m\), and \(m>0\) because their within-block
relations differ. Then \(G_+\geq mu^2\) and \(G_-\leq mv^2\), and
Proposition~\ref{prop:partition-exchange} gives the bound. Finally,
\(\|\mathbf D_{\mathcal Q}\|_2=\max_{C\in\mathcal Q}\|\Dff[C,C]\|_2\leq\max_C\tr(\Dff[C,C])\leq b\max_i[\Dff]_{ii}\).\hfill\(\square\)

\subsection{Proof of Theorem~\ref{thm:geometric-recovery}}
\label{app:recovery}
\paragraph{Recovery.}
By the definition of \(C_m\), the cell of \(z_1\) contains only points of
\(C_1\), that of \(z_M\) only points of \(C_{M-1}\), and that of \(z_{m+1}\)
only points of \(C_m\cup C_{m+1}\). By (i), the chain visits
\(z_1,\ldots,z_M\) in this order, so the neighbours of \(z_{m+1}\) are
\(z_m\) and \(z_{m+2}\). For \(x\in C_m\), \(z_m\) is one of its two nearest
inducing points and \(z_{m+2}\) is not, so
\(s_x=d_\theta^2(x,z_m)-d_\theta^2(x,z_{m+2})<0\); for \(y\in C_{m+1}\),
\(s_y>0\). Every cell therefore lists the points of the earlier group
first, each \(C_m\) is contiguous in \(\pi\), and, since \(|C_m|=b\), the cut
returns \(\mathcal Q\). A tie between a point's two adjacent anchors only
changes the cell that receives it, and ties within a group may be broken
arbitrarily.

\paragraph{Optimality.}
Write \(\rho(x)^2=k_\theta(x,x)\), \(c(x,y)=k_\theta(x,y)/(\rho(x)\rho(y))\), and let
\(k_+\) and \(k_-\) be the largest and smallest
\(k_\theta(x_i,x_i)\). Condition (ii) reads \(|c|\leq\delta\) outside the
groups and, since
\(c(x,y)=[\rho(x)^2+\rho(y)^2-d_\theta^2(x,y)]/(2\rho(x)\rho(y))\geq1-d_\theta^2(x,y)/(2\rho(x)\rho(y))\),
\(1-c(x,y)\leq h^2/(2k_-)\) inside a group. Exact conditioning gives
\(D(x,y)=\rho(x)\rho(y)D_c(x,y)\), where
\(D_c(x,y)=c(x,y)-c_{xZ}C_{ZZ}^{-1}c_{Zy}\) is the residual of \(c\) and
\(C_{ZZ}=[c(z_m,z_n)]\). Let
\(\eta=\max_m\sum_{n\neq m}|c(z_m,z_n)|\leq(M-1)\delta\), below one for \(\delta<1/(M-1)\), and
\(A=\max_i\sum_mc(x_i,z_m)^2\leq M\delta^2\). Then
\(\lambda_{\min}(C_{ZZ})\geq1-\eta\) and, by Cauchy--Schwarz,
\[
|c_{xZ}C_{ZZ}^{-1}c_{Zy}|\leq\zeta:=\frac{A}{1-\eta}\leq\frac{M\delta^2}{1-(M-1)\delta}.
\]
Hence \(D_c(x,x)\geq1-\zeta\), and \(|D_c(x,y)|\leq\delta+\zeta\) across
groups. Inside a group, the last inequality of
Section~\ref{sec:selection}, applied to \(c\), for which
\(d^2(x,y)=2-2c(x,y)\), gives
\(D_c(x,y)\geq\tfrac12[D_c(x,x)+D_c(y,y)]-(1-c(x,y))\geq1-\zeta-h^2/(2k_-)\).
Because \(h^2<2k_-\), the following choice has \(u>0\) for all
sufficiently small \(\delta\), as required by Lemma~\ref{lem:separation}:
\[
u=k_--\tfrac12h^2-k_-\zeta,\qquad
v=k_+(\delta+\zeta),\qquad
\kappa_{\mathcal Q}\leq1+\frac{bk_+}{\sigma^2},
\]
and \(\mathcal Q\) is the strict optimum among partitions into blocks of
size \(b\) whenever
\begin{equation}
k_--\tfrac12h^2-k_-\zeta
>\Bigl(1+\frac{bk_+}{\sigma^2}\Bigr)k_+(\delta+\zeta).
\label{eq:recovery-cert}
\end{equation}
As \(\delta\to0\), \(\zeta\to0\) and Eq.~\eqref{eq:recovery-cert} reduces to
\(h^2<2k_-\).\hfill\(\square\)

\paragraph{Kernels of the experiments.}
Place the inducing points on a line, consecutive ones at distance \(L\) in
lengthscale units, and every point of \(C_m\) within \(\varepsilon\) of the
midpoint of \(z_m\) and \(z_{m+1}\). Such a point is within
\(L/2+\varepsilon\) of \(z_m\) and \(z_{m+1}\) and farther than
\(3L/2-\varepsilon\) from the other inducing points.
\begin{itemize}
\item RBF and Mat\'ern-5/2: \(d_\theta\) increases with distance, so the
groups and (i) are as required for \(\varepsilon<L/2\), and (ii) holds with
\(h\) the kernel distance at \(2\varepsilon\), for which \(h^2<2\sigma_f^2\) (\(\sigma_f^2\) is the signal variance, part of \(\gamma\)),
and \(\delta\) the correlation at \(L/2-\varepsilon\), which vanishes as
\(L\) grows.
\item Amplitude-RBF, \(k(x,y)=a(x)a(y)k_{\mathrm{RBF}}(x,y)\) with
\(a(x)=\tfrac12+\operatorname{sigmoid}(w^\top x)\): the correlation \(c\),
hence (ii), is that of RBF; the distance conditions hold when \(a\) is
constant on the inducing points. Otherwise the nearest inducing point of a
distant point is the one of smallest amplitude, and they can fail.
\item Gibbs \citep{gibbs1997,paciorek2004}, with local lengthscales
\(\ell_k(x)=\ell_k[1+\operatorname{sigmoid}(w_k^\top x)]\in(\ell_k,\tau\ell_k)\), \(\tau=2\):
\[
c(x,y)=\prod_k\Bigl[\frac{2\ell_k(x)\ell_k(y)}{\ell_k(x)^2+\ell_k(y)^2}\Bigr]^{1/2}
\exp\Bigl[-\sum_k\frac{(x_k-y_k)^2}{\ell_k(x)^2+\ell_k(y)^2}\Bigr]
\]
lies between two Gaussian envelopes in the distance (e.g.\
\(c\leq e^{-r^2/8}\), \(r\) the distance in lengthscale units), which give
(ii) and the groups for large \(L\). For (i), with
\(\alpha_k^{(n)}=\ell_k(z_n)^2/\ell_k^2\in[1,\tau^2]\), the exponents of
\(c(z_m,z_{m+2})\) and \(c(z_m,z_{m+1})\) differ in the coordinate
along the inducing line by \(L^2\) times
\[
\frac{4}{\alpha^{(m)}+\alpha^{(m+2)}}-\frac{1}{\alpha^{(m)}+\alpha^{(m+1)}}
=\frac{3\alpha^{(m)}+4\alpha^{(m+1)}-\alpha^{(m+2)}}
{(\alpha^{(m)}+\alpha^{(m+2)})(\alpha^{(m)}+\alpha^{(m+1)})}
\geq\frac{7-\tau^2}{4\tau^4},
\]
and every still farther \(z_j\), \(j>m+2\), has a squared separation at
least \(4L^2\), giving the same positive lower bound on the exponent
gap. The Gibbs prefactors are bounded above and below by constants
depending only on \(\tau\), so the positive exponent gap dominates for
large \(L\). Thus \(z_{m+1}\) is the nearest to \(z_m\) whenever
\(\tau<\sqrt7\). For \(\tau>\sqrt7\), the lengthscales
\(\ell(z_m)=\ell(z_{m+1})=\ell\), \(\ell(z_{m+2})=\tau\ell\) make the chain
skip \(z_{m+1}\) for sufficiently large \(L\).
\end{itemize}
The argument does not depend on the input dimension, and the strict
margins persist under small perturbations of the inputs.

\paragraph{An explicit RBF family with small blocks.}
The sufficient regime need not disappear when both the block and inducing
fractions vanish. For any integer \(M\geq3\), set \(b=M\),
\(N=M(M-1)\), unit signal and noise variances, and unit RBF lengthscale.
Put \(z_j=(j-1)L\) on the real line and \(b\) labelled observations at
each midpoint \((j-\tfrac12)L\), \(j=1,\ldots,M-1\), where
\(L=\sqrt{8\log(20M^2)}\). The adjacent observation--anchor correlation
is \(\delta=1/(20M^2)\); every other observation--anchor correlation is
smaller, while cross-group observation and distinct inducing correlations
are at most \(\delta^4\). The within-group distance is \(h=0\), and the
nearest-anchor chain follows the line. Consequently
\[
\zeta=\frac{M\delta^2}{1-(M-1)\delta}
\leq\frac{1}{380M^3},\qquad
u=1-\zeta,\quad v=\delta+\zeta,\quad
\kappa_{\mathcal Q}\leq1+M.
\]
Here \(u>1/2\) and \(\kappa_{\mathcal Q}v<1/2\), so
Eq.~\eqref{eq:recovery-cert} holds and the selected partition strictly
outperforms every other balanced partition at fixed state. Repeated
observations only simplify the construction: sufficiently small distinct
perturbations preserve its strict inequalities. Meanwhile,
\(b/N=M/N=1/(M-1)\to0\). Thus the strict advantage over a uniform random
partition can persist with sublinear blocks and inducing sets; this is a
non-vacuity example, not a claim that the experimental data satisfy the
uniform-separation condition.

\paragraph{Numerical check.}
The example of Figure~\ref{fig:geometric-recovery-main} (RBF, two input
dimensions, \(N=48\), \(M=5\), \(b=12\), unit lengthscales,
\(\sigma_f^2=\sigma^2=1\)) is recovered by the implementation and satisfies
Eq.~\eqref{eq:recovery-cert} with margin \(0.197\); so does a version
perturbed in three extra coordinates. With four inducing points on a line,
groups of four points, \(\ell=\sigma_f^2=1\) and \(\sigma^2=0.1\), we
enumerated all \(5{,}775\) balanced partitions. For RBF and for Gibbs with
several warps, Algorithm~\ref{alg:sbt} returned \(\mathcal Q\), the unique
optimum, at every spacing tried, including spacings well below those at
which Eq.~\eqref{eq:recovery-cert} holds. For Amplitude-RBF with a slope
along the line, \(\mathcal Q\) remained the unique optimum while most points
fell in a non-adjacent cell at large spacing; with the amplitude constant on
the inducing points, recovery held at every spacing.

\section{Block-Diagonal Predictor}
\label{app:hard-block-predictor}
This appendix specifies the predictor of Section~\ref{sec:bounds} for given
parameters \(\theta\), \(\mathbf Z\), \(q(\mathbf u)=\N(\mathbf m,\mathbf S)\) and a
given partition \(\{\mathcal B_r\}\) of the training points;
Appendix~\ref{app:prediction-partitions} states which partition each method uses.

\paragraph{Definition.}
Let \(\mathbf D_r=\Dff[\mathcal B_r,\mathcal B_r]\) and
\(\mathbf d_{\ast r}=\mathbf k_{\ast\mathcal B_r}
-\mathbf k_{\ast\mathbf u}\Kuu^{-1}\mathbf K_{\mathbf u\mathcal B_r}\).
Each block is represented by its training input closest to the block mean,
after scaling the coordinates by the lengthscales, and a test input
\(x_\ast\) is assigned to the block \(r\) whose representative \(x_j\)
minimizes the kernel distance \(d_\theta^2(x_\ast,x_j)\) of
Eq.~\eqref{eq:anchor-distance}, with ties broken by block index; no test
response is used. The predictor keeps the mean \(\mu_\ast=\mathbf A_\ast\mathbf m\)
and has latent variance \(v_\ast=v_\ast^{\mathrm{SVGP}}-c_\ast\), where
\begin{equation}
c_\ast=\mathbf d_{\ast r}\mathbf V_r\mathbf d_{r\ast},\qquad
\mathbf V_r=
\begin{cases}
\mathbf D_r^{\dagger/2}\boldsymbol\Lambda_r\mathbf D_r^{\dagger/2}, & \text{T-SVGP},\\[1mm]
(\mathbf D_r+\sigma^2\Id)^{-1}, & \text{BT-SVGP},
\end{cases}
\label{eq:hard-block-correction}
\end{equation}
\(\boldsymbol\Lambda_r\) is the diagonal matrix of Section~\ref{sec:bounds}, and
\(\mathbf D_r^{\dagger/2}\) is the symmetric square root of the pseudoinverse
\(\mathbf D_r^\dagger\), equal to \(\mathbf D_r^{-1/2}\) when \(\mathbf D_r\succ0\).
The same formulas give the T-SGPR and BT-SGPR predictors, and SGPR and SVGP
have \(c_\ast=0\). The observation variance is \(v_\ast+\sigma^2\). Since the mean
is unchanged, the predictor affects NLL but not RMSE. It is a finite
block-conditional approximation, not the dense variational covariance of
Eq.~\eqref{eq:bui-predictive-variance}; Appendix~\ref{app:predictive-residual-check}
compares the two.

\paragraph{Positivity.}
No variance is clipped, because \(v_\ast\geq0\) by construction. If
\(\mathbf D_r\succ0\), then \((\mathbf D_r+\sigma^2\Id)^{-1}\preceq\mathbf D_r^{-1}\), and
\(0\preceq\boldsymbol\Lambda_r\preceq\Id\) gives
\(\mathbf D_r^{-1/2}\boldsymbol\Lambda_r\mathbf D_r^{-1/2}\preceq\mathbf D_r^{-1}\), so
\(0\preceq\mathbf V_r\preceq\mathbf D_r^{-1}\) in both cases. The residual covariance
of the block and the test input,
\(\bigl[\begin{smallmatrix}\mathbf D_r&\mathbf d_{r\ast}\\ \mathbf d_{\ast r}&d_{\ast\ast}\end{smallmatrix}\bigr]\),
is a Schur complement of a kernel Gram matrix with \(\Kuu\succ0\) and hence
positive semidefinite. Therefore
\(c_\ast\leq\mathbf d_{\ast r}\mathbf D_r^{-1}\mathbf d_{r\ast}\leq d_{\ast\ast}\) and
\(v_\ast\geq\mathbf A_\ast\mathbf S\mathbf A_\ast^\top\geq0\). If \(\mathbf D_r\) is
singular, the same positive semidefiniteness places \(\mathbf d_{r\ast}\) in the
range of \(\mathbf D_r\), on which both choices of \(\mathbf V_r\) are dominated by
\(\mathbf D_r^\dagger\), and the argument holds with \(\mathbf D_r^\dagger\) in place
of \(\mathbf D_r^{-1}\). With \(\sigma^2>0\), the observation variance is strictly
positive. The argument concerns this single-block predictor, not predictors
that combine several blocks.

\paragraph{Numerical implementation.}
All computations use double precision. For BT, \(\mathbf V_r\) is applied
through the Cholesky factor of \(\mathbf D_r+\sigma^2\Id\), which is positive
definite for \(\sigma^2>0\); a jitter of \(10^{-8}\max(1,\sigma^2)\) is added only
if the factorization fails. For T, \(\mathbf D_r=\mathbf W\operatorname{diag}(\lambda_i)\mathbf W^\top\)
is diagonalized, and eigenvalues
\(\lambda_i\leq\tau_r=10\,\epsilon\,|\mathcal B_r|\,\lambda_{\max}(\mathbf D_r)\), with
\(\epsilon\) the machine epsilon, receive weight zero, which implements
\(\mathbf D_r^\dagger\); the block is rejected if
\(\lambda_{\min}(\mathbf D_r)<-\max(\tau_r,10^{-10})\). We record the fraction
\(\sum_{\lambda_i\leq\tau_r}\alpha_i^2/\sum_i\alpha_i^2\) of \(\mathbf d_{r\ast}\) on the
truncated eigenvectors, where \(\alpha_i=\mathbf w_i^\top\mathbf d_{r\ast}\). A
prediction is rejected if a variance is non-finite or non-positive, or if
\(d_{\ast\ast}-c_\ast<-10^{-7}\). Appendix~\ref{app:evaluation} reports the
observed values.

\section{Split-Batch Estimator}
\label{app:svgp}
For a partition \(\{\mathcal B_r\}_{r=1}^B\), \(B=\lceil N/b\rceil\), the block gains
\[
g_r=\sum_{i\in\mathcal B_r}\log\Bigl(1+\frac{[\Dff]_{ii}}{\sigma^2}\Bigr)
-\log\bigl|\Id+\sigma^{-2}\Dff[\mathcal B_r,\mathcal B_r]\bigr|
\]
are nonnegative by Hadamard's inequality, and
\(\mathcal L_{\mathrm{BT}}=\mathcal L_{\mathrm T}+\tfrac12\sum_rg_r\). For a
likelihood batch \(\mathcal B\subset\{1,\ldots,N\}\), drawn independently of
the blocks, and blocks \(r_1,\ldots,r_K\),
\begin{equation}
\widehat{\mathcal L}_{\mathrm{split}}
=\widehat{\mathcal L}_{\mathrm T}(\mathcal B)+\frac{B}{2K}\sum_{k=1}^Kg_{r_k},
\label{eq:split}
\end{equation}
\[
\widehat{\mathcal L}_{\mathrm T}(\mathcal B)=-\KL[q(\mathbf u)\Vert p(\mathbf u)]
+\frac{N}{|\mathcal B|}\sum_{i\in\mathcal B}\Bigl\{\E_{q(\mathbf u)}\log\N(y_i;\mathbf k_{i\mathbf u}\Kuu^{-1}\mathbf u,\sigma^2)
-\tfrac12\log\Bigl(1+\frac{[\Dff]_{ii}}{\sigma^2}\Bigr)\Bigr\}.
\]
If \(\mathcal B\) is a uniform subset and the \(r_k\) are uniform on
\(\{1,\ldots,B\}\), then \(\E\widehat{\mathcal L}_{\mathrm T}(\mathcal B)=\mathcal L_{\mathrm T}\)
and \(\E\,\tfrac BK\sum_kg_{r_k}=\sum_rg_r\), so
\(\E\widehat{\mathcal L}_{\mathrm{split}}=\mathcal L_{\mathrm{BT}}\) at fixed
parameters and partition. The experiments traverse batches and blocks in
separate cycles without replacement, with uniform weights; the statement
above does not cover gradients along such a cycle.

\paragraph{Cost.}
After the \(O(M^3)\) Cholesky factorization of \(\Kuu\), a likelihood point
needs only \(\mathbf k_{i\mathbf u}\Kuu^{-1}\) and \([\Dff]_{ii}\), at cost
\(O(M^2)\). A block needs its cross-covariances, its residual
\(\Dff[\mathcal B_r,\mathcal B_r]\) and the Cholesky factor of
\(\Id+\sigma^{-2}\Dff[\mathcal B_r,\mathcal B_r]\), at cost
\(O(bM^2+b^2M+b^3)\). Counting each term with unit constant, an update costs
\[
C_{\mathrm{split}}=C_{\mathrm{shared}}+|\mathcal B|M^2,\qquad
C_{\mathrm{shared}}=M^3+K(bM^2+b^2M+b^3),
\]
where \(C_{\mathrm{shared}}\) is the cost when the \(K\) blocks also serve
as the likelihood batch. A shared batch twice as large costs
\(C_{\mathrm{double}}=M^3+2K(bM^2+b^2M+b^3)\approx2C_{\mathrm{shared}}\),
because the block work, Cholesky factorizations included, doubles. For \(|\mathcal B|=Kb\),
\[
1\leq\frac{C_{\mathrm{split}}}{C_{\mathrm{shared}}}
\leq1+\frac{M^2}{M^2+bM+b^2},
\]
which is close to one for \(M\leq b\). In our runs \(|\mathcal B|=1000\),
\(K=2\) and \(b=500\), so \(|\mathcal B|=Kb\). Under the leading-term operation-count model with unit
coefficients, \(C_{\mathrm{split}}/C_{\mathrm{shared}}\simeq1.05\) for \(M=128\),
\(|\mathcal B|=1000\), \(K=2\) and \(b=500\) (the diagnostic below), while a
shared batch of 2000 points would give \(1.99\). This is an operation-count
estimate, not a measured runtime overhead. For \(M=64,256,512\) the factor is
\(1.01\), \(1.14\) and \(1.29\), against \(2.00\), \(1.96\) and \(1.85\) for the
doubled shared batch. The update is \(O(M^3)\) for \(|\mathcal B|,b=O(M)\)
and fixed \(K\).

\begin{figure}[H]
\centering
\includegraphics[width=.48\textwidth]{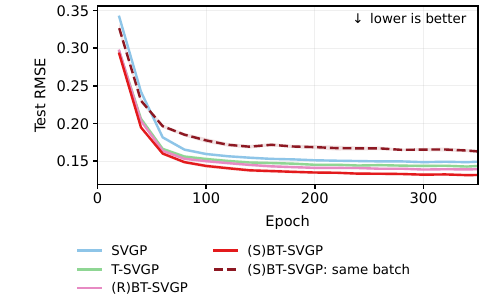}\hfill
\includegraphics[width=.48\textwidth]{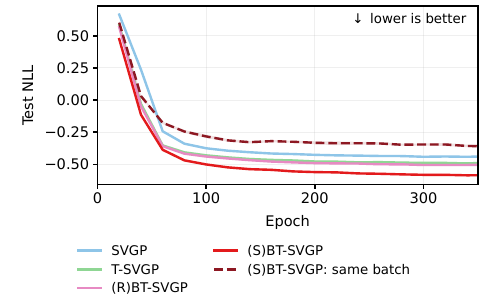}
\caption{Test RMSE (left) and NLL (right) of the split-batch diagnostic of
Figure~\ref{fig:svgp-batch-ablation}, with the block-diagonal predictor at the
saved parameters; dashed: shared batch. Mean and one standard error over ten
seeds.}
\label{fig:svgp-batch-test}
\end{figure}

The diagnostic of Figures~\ref{fig:svgp-batch-ablation}
and~\ref{fig:svgp-batch-test} uses RBF--Pol, \(M=128\) and ten seeds per
curve, trained for 350 epochs with Lion \citep{chen2023lion} at learning rate
\(5\times10^{-3}\) for \(\theta\) and \(\mathbf Z\) and natural gradients for
\(q(\mathbf u)\) with step 0.05; the dashed (S)BT-SVGP curve uses one shared
batch for both terms. Parameters are saved every 20 epochs and at epoch 350,
and the test RMSE and NLL are computed afterwards at these snapshots with the
block-diagonal predictor, so every point of Figure~\ref{fig:svgp-batch-test}
uses the predictor of the main results.

\section{Experimental Protocol}
\label{app:experiments}
This appendix describes the training runs (Appendix~\ref{app:training-protocols}),
the evaluation (Appendix~\ref{app:evaluation}), the partitions used for prediction
(Appendix~\ref{app:prediction-partitions}) and the comparison with the dense
predictor (Appendix~\ref{app:predictive-residual-check}). The code and aggregate
results are available at
\url{https://github.com/Jollikop/geometry-aware-block-selection}; the datasets are cited in
Section~\ref{sec:experiments} and Appendix~\ref{app:svgp-frozen}.

\subsection{Training}
\label{app:training-protocols}
\paragraph{Selector.}
The implementation starts the chain at the first inducing point, breaks
nearest-anchor and chain ties by the smallest anchor index, and sorts points
stably by chain position, boundary score and distance to their anchor. Empty
cells keep their anchors in the chain, and a final short block is retained
when \(b\) does not divide \(N\).

\paragraph{Snelson.}
\(N=200\), \(M=5\), blocks of 20 points and at most 60 L-BFGS iterations;
the (S)BT-SGPR partition is selected at initialization and reselected after
iterations 2, 5, 10 and 30 (Figure~\ref{fig:snelson}). The Snelson setup follows
the public example code of \citet{bui2025tight}
(\url{https://github.com/thangbui/tighter_sparse_gp}).

\paragraph{Collapsed SGPR.}
Kin40k with 5000 training and 2000 test points, ten seeds,
\(M\in\{128,256,512\}\) and \(b=500\). L-BFGS-B runs for at most 100000
iterations; a fit is accepted when the per-datum objective varies by less
than \(10^{-4}\) over two consecutive 100-step windows and the largest
absolute per-datum gradient is below \(10^{-3}\), or when L-BFGS-B stops
at a stationary point and the remaining trajectory and gradient meet the
same tolerances. \(\mathbf Z\) is initialized by greedy variance selection
\citep{burt2020convergence}, followed by a short variational-EM warm-up.
(S)BT-SGPR and (METIS)BT-SGPR reselect the partition after iterations 15, 30,
50 and 75 and keep it fixed thereafter; (R)BT-SGPR keeps its random
partition. The refined sparse predictor fixes the kernel and noise
parameters and optimizes \(\mathbf Z\); four of 30 SGPR self-transplant
controls, in which an SGPR donor is refined under its own objective and
should not change, exceed the invariance thresholds (\(5\times10^{-4}\) for the
per-datum ELBO, \(10^{-3}\) for the sparse NLL) and are kept.
Table~\ref{tab:kin40k-sgpr-refined} reports \(M=512\). The search times in
Table~\ref{tab:kin40k-sgpr} are warm partition-build timings from the
training runs, averaged over ten seeds.

\begin{table}[!ht]
\caption{Sparse SGPR refinement of the same \(M=512\) donors.
Kernel and noise parameters are fixed; inducing locations are optimized.
Direct sparsification drops the covariance correction without refinement,
leaving the predictive mean and RMSE unchanged.
Mean (SE in the last decimal places), ten seeds; bold: lowest mean in each column.}
\label{tab:kin40k-sgpr-refined}
\centering\footnotesize
\setlength{\tabcolsep}{3pt}
\begin{tabular}{@{}lrrrr@{}}
\toprule
& \multicolumn{2}{c}{Direct sparse} & \multicolumn{2}{c}{Refined sparse} \\
\cmidrule(lr){2-3}\cmidrule(lr){4-5}
Donor & RMSE & NLL & RMSE & NLL \\
\midrule
SGPR & .229(2) & $-$.041(7) & .229(2) & $-$.041(7) \\
T-SGPR & .205(2) & $-$.122(7) & .207(2) & $-$.119(7) \\
(R)BT-SGPR & .197(2) & $-$.144(7) & .200(2) & $-$.142(7) \\
(S)BT-SGPR & \textbf{.185(2)} & \textbf{$-$.158(7)} & \textbf{.190(2)} & \textbf{$-$.164(7)} \\
\bottomrule
\end{tabular}
\end{table}

\paragraph{Stochastic SVGP.}
Each seed has its own 90/10 split; inputs and targets are standardized
with training statistics. Donors use Adam \citep{kingma2015adam} with
learning rate \(5\times10^{-3}\) for \(\theta\) and \(\mathbf Z\), natural gradients
\citep{salimbeni2018natural} for \(q(\mathbf u)\) with step 0.05, 1000 likelihood
points and two blocks of 500 points per step, for 350 epochs (RBF,
Mat\'ern-5/2), 500 epochs (Amplitude-RBF), or 300 epochs from the RBF
checkpoint (Gibbs). Selected partitions are refreshed every 40 epochs
(epochs 40--160 for RBF and Mat\'ern-5/2, 40--480 for Amplitude-RBF, 40 and 80
for Gibbs), and training continues with the last refreshed partition;
(R)BT-SVGP keeps its random partition. The transplant freezes \(\theta\), also
freezes \(\mathbf Z\) for the first 10 epochs, updates \(\mathbf Z\) with Adam at
learning rate \(10^{-3}\) and \(q(\mathbf u)\) with natural gradients of step 0.05,
and resets Adam at epoch 150 of 300. The training times in
Section~\ref{sec:svgp-experiments} come from these runs; for RBF, eight of the
ten seeds ran on an NVIDIA A100-PCIE-40GB and two on an NVIDIA H100 80GB.

\subsection{Evaluation}
\label{app:evaluation}
Test metrics are computed after training from the saved parameters, without
retraining, and each record is checked against its saved inputs and an
evaluator hash before aggregation. On the standardized scale, the test NLL
of a seed is
\[
\frac1{N_\ast}\sum_{\ast}\Bigl[\tfrac12\log\bigl(2\pi(v_\ast+\sigma^2)\bigr)
+\frac{(y_\ast-\mu_\ast)^2}{2(v_\ast+\sigma^2)}\Bigr],
\]
where \(\mu_\ast\) and \(v_\ast+\sigma^2\) are the predictive mean and observation
variance; figures and tables report the mean and one standard error over
seeds. The block-diagonal predictor of Appendix~\ref{app:hard-block-predictor} is
evaluated for the Snelson fits, for the RBF--Kin40k SGPR methods at \(M=512\)
(40 records) and for T-SVGP, (R)BT-SVGP and (S)BT-SVGP over all kernels,
datasets, inducing budgets and seeds (3840 records), on NVIDIA A100 PCIe GPUs
(40GB, and 80GB for 60 records). None of the 3880 records was rejected or
needed jitter. The smallest \(d_{\ast\ast}-c_\ast\) was \(1.1\times10^{-8}\) for SVGP
and \(1.1\times10^{-3}\) for SGPR, the smallest eigenvalue of a T block was
\(-1.9\times10^{-12}\), and the largest truncated fraction was
\(2.5\times10^{-7}\).

\subsection{Prediction Partitions}
\label{app:prediction-partitions}
We write \(\mathcal P_{\mathrm{train}}\) for the partition that defines the
training objective at the last optimization step and \(\mathcal P_{\mathrm{pred}}\)
for the partition given to the predictor of
Appendix~\ref{app:hard-block-predictor}. The SGPR partitions are fixed after the
last reselection, and (R)BT-SVGP keeps its random partition throughout, so
for these methods \(\mathcal P_{\mathrm{pred}}=\mathcal P_{\mathrm{train}}\). T-SGPR and
T-SVGP have no training partition; their \(\mathcal P_{\mathrm{pred}}\) is built by
Algorithm~\ref{alg:sbt} at the saved parameters, with the block size of the BT
methods. For (S)BT-SVGP, \(\theta\), \(\mathbf Z\) and \(q(\mathbf u)\) keep changing
after the last refresh, so \(\mathcal P_{\mathrm{pred}}\) is rebuilt by
Algorithm~\ref{alg:sbt} at the saved parameters and need not equal
\(\mathcal P_{\mathrm{train}}\). The rebuild changes neither the parameters nor the
reported final objective, which uses \(\mathcal P_{\mathrm{train}}\); the predictor
uses the optimal BT scaling \(\mathbf M_r^\star=(\Id+\sigma^{-2}\mathbf D_r)^{-1}\) of
each block of \(\mathcal P_{\mathrm{pred}}\), which gives \(\mathbf V_r\) in
Eq.~\eqref{eq:hard-block-correction}.

\paragraph{Training versus prediction partition.}
Since \(c_\ast\) depends on \(\mathcal P_{\mathrm{pred}}\), a block-diagonal NLL
reflects both the fitted state \((\theta,\mathbf Z,q(\mathbf u))\) and the prediction
partition: T-SVGP and (S)BT-SVGP predict with blocks from
Algorithm~\ref{alg:sbt}, (R)BT-SVGP with random blocks of the same size. The
NLL comparisons in Figure~\ref{fig:svgp-final} and Appendix~\ref{app:svgp-frozen}
therefore compare complete pipelines. To separate the two effects on the
configurations of Figure~\ref{fig:svgp-final}, we kept the fitted (R)BT-SVGP and
(S)BT-SVGP models fixed and predicted with the same kind of partition for
both: blocks from Algorithm~\ref{alg:sbt} at the saved parameters, or uniformly
random blocks of the same size (Table~\ref{tab:partition-controls}; 640
records, none rejected). Under both common prediction-partition rules, (S)BT-SVGP has
lower mean NLL than (R)BT-SVGP in all 32 configurations (by 0.012 to 0.083,
each by more than two standard errors, under the common selected rule; by
0.005 to 0.042 under the common random rule). Under the common selected
rule, 60--97\% (mean 77\%) of the original NLL gap of
Figure~\ref{fig:svgp-final} remains. Thus the improvement is not explained
solely by assigning selected prediction blocks to one donor and random blocks
to the other. Its magnitude depends on the prediction rule: selected
prediction blocks lower the NLL of both models, more for (S)BT-SVGP (by 0.006
to 0.079) than for (R)BT-SVGP (by 0.001 to 0.034). These controls do not
uniquely apportion the gain between the fitted state and the prediction
partition.

\begin{table}[!ht]
\caption{Prediction-partition control on the configurations of Figure~\ref{fig:svgp-final}: test NLL of (S)BT-SVGP minus that of (R)BT-SVGP (negative favours (S)BT-SVGP), mean (SE) over ten paired seeds. Current: each donor with its own \(\mathcal P_{\mathrm{pred}}\), as in Figure~\ref{fig:svgp-final}. Common selected rule: both donors with \(\mathcal P_{\mathrm{pred}}\) from Algorithm~\ref{alg:sbt} at their saved parameters. Common random rule: both donors with a uniformly random partition into blocks of 500. The common selected rule applies Algorithm~\ref{alg:sbt} separately at each donor's saved parameters; block memberships need not coincide. Under the common random rule the donors do not share an index permutation: (R)BT-SVGP keeps the random partition it was trained with, and (S)BT-SVGP receives an independent uniformly random permutation in each paired seed. Donor parameters are frozen; RMSE is unchanged.}
\label{tab:partition-controls}
\centering\footnotesize
\setlength{\tabcolsep}{4pt}
\begin{tabular}{@{}llrrrr@{}}
\toprule
Kernel & Dataset & \(M\) & Current & \begin{tabular}[b]{@{}r@{}}Common\\selected rule\end{tabular} & \begin{tabular}[b]{@{}r@{}}Common\\random rule\end{tabular} \\
\midrule
RBF & Pol & 64 & -.059(5) & -.036(5) & -.017(4) \\
RBF & Pol & 128 & -.073(6) & -.045(6) & -.014(3) \\
RBF & Pol & 256 & -.078(4) & -.047(4) & -.011(4) \\
RBF & Pol & 512 & -.084(3) & -.051(4) & -.013(2) \\
\addlinespace[1pt]
RBF & Supercond. & 64 & -.033(2) & -.023(2) & -.008(3) \\
RBF & Supercond. & 128 & -.035(2) & -.026(2) & -.005(3) \\
RBF & Supercond. & 256 & -.033(2) & -.025(2) & -.006(3) \\
RBF & Supercond. & 512 & -.029(1) & -.023(1) & -.008(2) \\
\addlinespace[1pt]
RBF & Kin40k & 64 & -.047(3) & -.037(3) & -.028(3) \\
RBF & Kin40k & 128 & -.075(4) & -.053(4) & -.031(3) \\
RBF & Kin40k & 256 & -.103(4) & -.072(4) & -.035(3) \\
RBF & Kin40k & 512 & -.117(3) & -.083(3) & -.038(2) \\
\addlinespace[1pt]
RBF & Protein & 64 & -.015(2) & -.012(2) & -.006(2) \\
RBF & Protein & 128 & -.028(2) & -.023(2) & -.013(1) \\
RBF & Protein & 256 & -.026(1) & -.021(1) & -.011(1) \\
RBF & Protein & 512 & -.022(1) & -.017(1) & -.009(1) \\
\addlinespace[1pt]
Amplitude-RBF & Pol & 64 & -.066(11) & -.053(10) & -.036(9) \\
Amplitude-RBF & Pol & 128 & -.058(4) & -.041(4) & -.025(2) \\
Amplitude-RBF & Pol & 256 & -.060(4) & -.045(4) & -.025(4) \\
Amplitude-RBF & Pol & 512 & -.047(4) & -.034(3) & -.021(3) \\
\addlinespace[1pt]
Amplitude-RBF & Supercond. & 64 & -.033(2) & -.031(2) & -.024(2) \\
Amplitude-RBF & Supercond. & 128 & -.041(1) & -.038(2) & -.032(3) \\
Amplitude-RBF & Supercond. & 256 & -.043(2) & -.041(2) & -.032(2) \\
Amplitude-RBF & Supercond. & 512 & -.028(1) & -.027(2) & -.022(2) \\
\addlinespace[1pt]
Amplitude-RBF & Kin40k & 64 & -.056(4) & -.047(4) & -.034(4) \\
Amplitude-RBF & Kin40k & 128 & -.081(4) & -.058(4) & -.031(3) \\
Amplitude-RBF & Kin40k & 256 & -.102(2) & -.072(2) & -.036(2) \\
Amplitude-RBF & Kin40k & 512 & -.112(2) & -.082(2) & -.042(2) \\
\addlinespace[1pt]
Amplitude-RBF & Protein & 64 & -.017(1) & -.015(1) & -.010(2) \\
Amplitude-RBF & Protein & 128 & -.032(1) & -.029(1) & -.021(1) \\
Amplitude-RBF & Protein & 256 & -.022(1) & -.019(1) & -.013(1) \\
Amplitude-RBF & Protein & 512 & -.017(1) & -.015(1) & -.011(1) \\
\bottomrule
\end{tabular}
\end{table}

\subsection{Block-Diagonal versus Dense Prediction}
\label{app:predictive-residual-check}
\begin{figure}[!htbp]
\centering
\includegraphics[width=.88\textwidth]{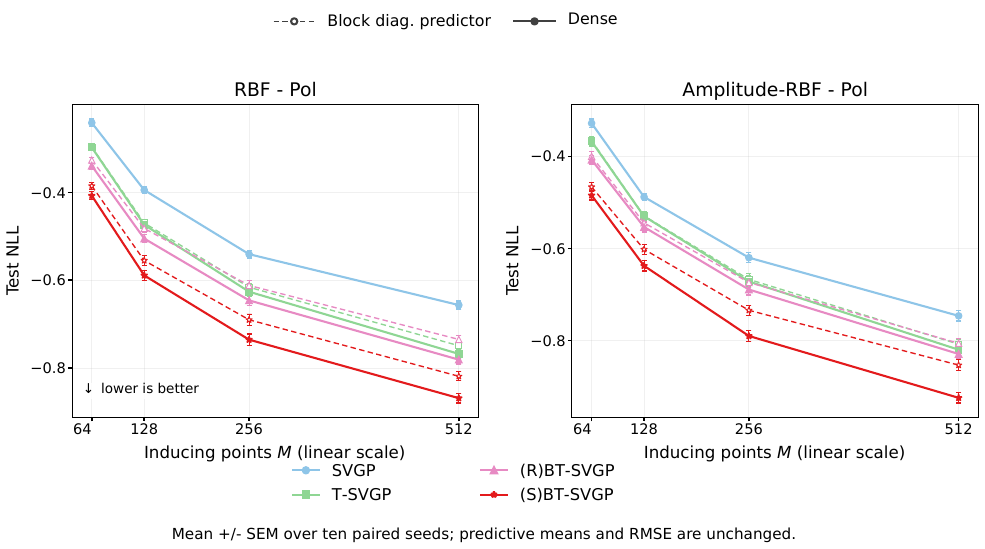}
\caption{Pol test NLL for the block-diagonal predictor used in the main results
(dashed) and the dense variational predictor (solid), paired at
the same saved parameters.
Mean and one standard error over ten paired seeds.}
\label{fig:pol-real-predictor-nll}
\end{figure}

We compare the block-diagonal predictor with the dense variational predictor
of Eq.~\eqref{eq:bui-predictive-variance}, which uses the full \(\Dff\), at the
same saved parameters: RBF--Kin40k SGPR at \(M=512\)
(Table~\ref{tab:kin40k-sgpr-real-nll}) and RBF and Amplitude-RBF SVGP on Pol
(Figure~\ref{fig:pol-real-predictor-nll}), with 280 pairs of matching
checkpoints and ten paired seeds per configuration. For the BT methods, the
dense scaling is built from the same \(\mathcal P_{\mathrm{pred}}\). The dense
predictor needs \(\Dff^{-1/2}\): with
\(\Dff=\mathbf W\operatorname{diag}(\lambda_i)\mathbf W^\top\), eigenvalues
\(\lambda_i\leq\tau=10\,\epsilon N\lambda_{\max}(\Dff)\) receive weight zero. None
falls below \(\tau\) on Kin40k, whereas up to 2791 of the 13500 eigenvalues do
on Pol. The fraction of \(\mathbf d_{\mathbf f\ast}\) on these eigenvectors is below
\(6\times10^{-11}\), and the smallest dense latent variance is \(7\times10^{-5}\).
Since \(\Dff^{-1/2}\) weights each coordinate by \(\alpha_i/\sqrt{\lambda_i}\), with
\(\alpha_i=\mathbf w_i^\top\mathbf d_{\mathbf f\ast}\), a small fraction does not by itself
bound the error, so we repeated the 280 dense evaluations with the factor 10
in \(\tau\) replaced by 1, \(10^2\), \(10^4\) and \(10^6\) (1400 evaluations, none
rejected; factor 10 reproduces the stored NLL to \(4\times10^{-16}\)). On Pol, where
up to 2023 eigenvalues are discarded at factor 1 and up to 10885 at \(10^6\),
the test observation variances change by at most \(7\times10^{-4}\) in relative
terms for factors 1 and \(10^2\) and by at most \(3.3\%\) at \(10^6\), and the mean
test NLL by at most \(7.3\times10^{-5}\); on Kin40k nothing changes. The dense NLL
is lower than the block-diagonal NLL in every displayed configuration, by at
most 0.054 for SGPR and 0.071 for SVGP, but the SGPR ordering and the leading
SVGP method on Pol are unchanged; means and RMSE are identical. These are
empirical checks on the displayed data, not a bound on the block-diagonal
approximation.

\begin{table}[!ht]
\caption{RBF--Kin40k, $M=512$: test NLL (mean (SE) over ten paired seeds; lower is better). Block diag. is the predictor in the main results; dense uses the full $\mathbf D_{\mathbf{ff}}$ in Eq.~\eqref{eq:bui-predictive-variance}. Bold marks the best of the four principal methods; $^\dagger$ denotes comparators.}
\label{tab:kin40k-sgpr-real-nll}
\centering\footnotesize
\setlength{\tabcolsep}{3pt}
\begin{tabular}{@{}lrr@{}}
\toprule
Method & Block diag. NLL & Dense NLL \\
\midrule
SGPR & -.041(7) & -.041(7) \\
T-SGPR & -.146(7) & -.160(8) \\
(R)BT-SGPR & -.179(7) & -.213(8) \\
(S)BT-SGPR & \textbf{-.240(7)} & \textbf{-.294(7)} \\
\midrule
(METIS)BT-SGPR$^{\dagger}$ & -.296(6) & -.336(7) \\
\midrule
Exact GP$^{\dagger}$ & --- & -.618(14) \\
\bottomrule
\end{tabular}
\end{table}

\FloatBarrier
\section{Full SVGP Campaign}
\label{app:svgp-frozen}
\begin{landscape}
\begin{figure}[p]
\centering
\includegraphics[width=.99\linewidth,height=.83\textheight,keepaspectratio]{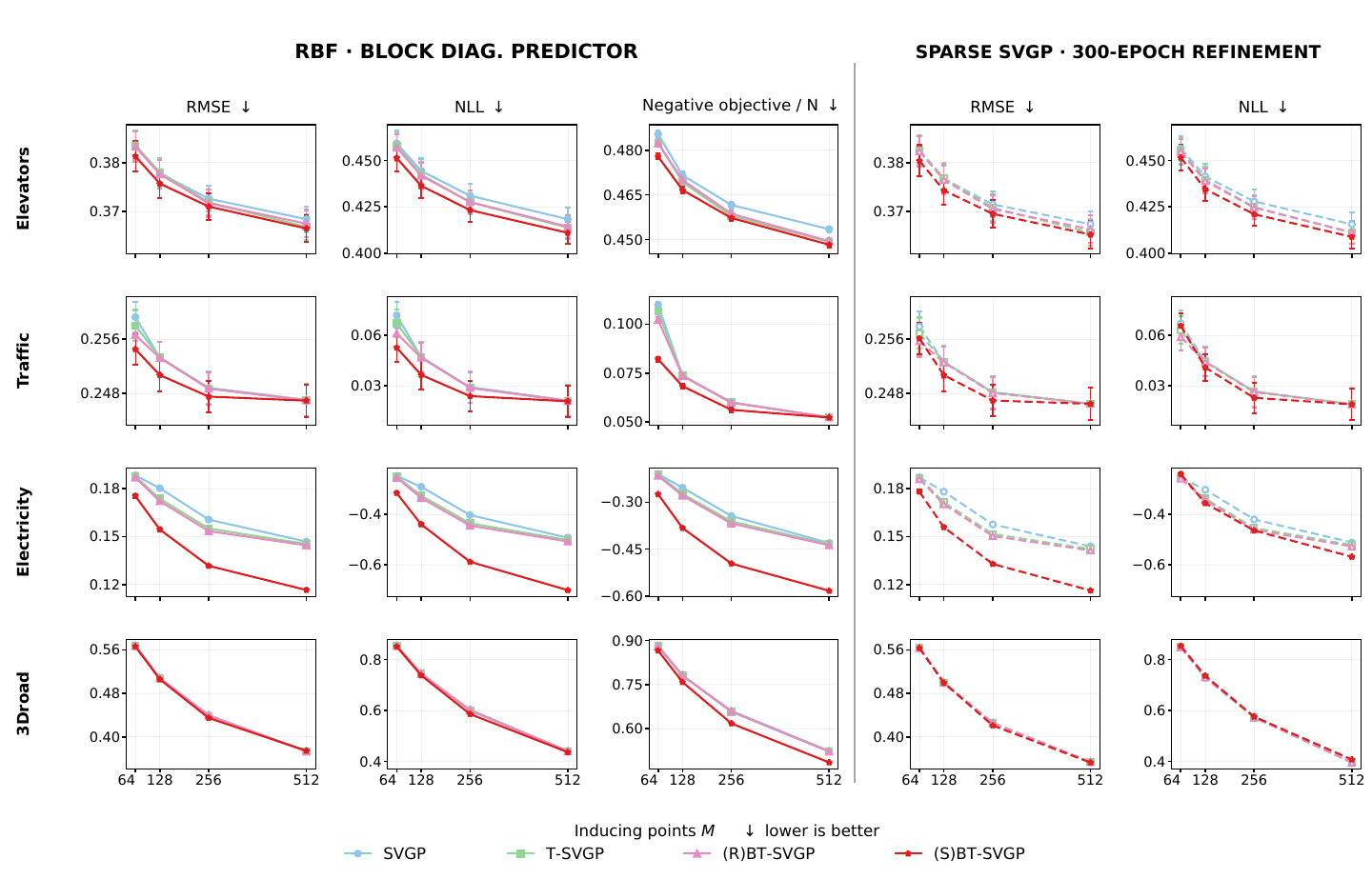}
\caption{RBF SVGP campaign: Elevators, Traffic, Electricity, 3Droad. Left: block-diagonal predictor and negative final objective (solid); right: sparse predictor after the transplant (dashed). Mean and one standard error over ten seeds.}
\label{fig:campaign-rbf-additional}
\end{figure}
\end{landscape}

\begin{landscape}
\begin{figure}[p]
\centering
\includegraphics[width=.99\linewidth,height=.83\textheight,keepaspectratio]{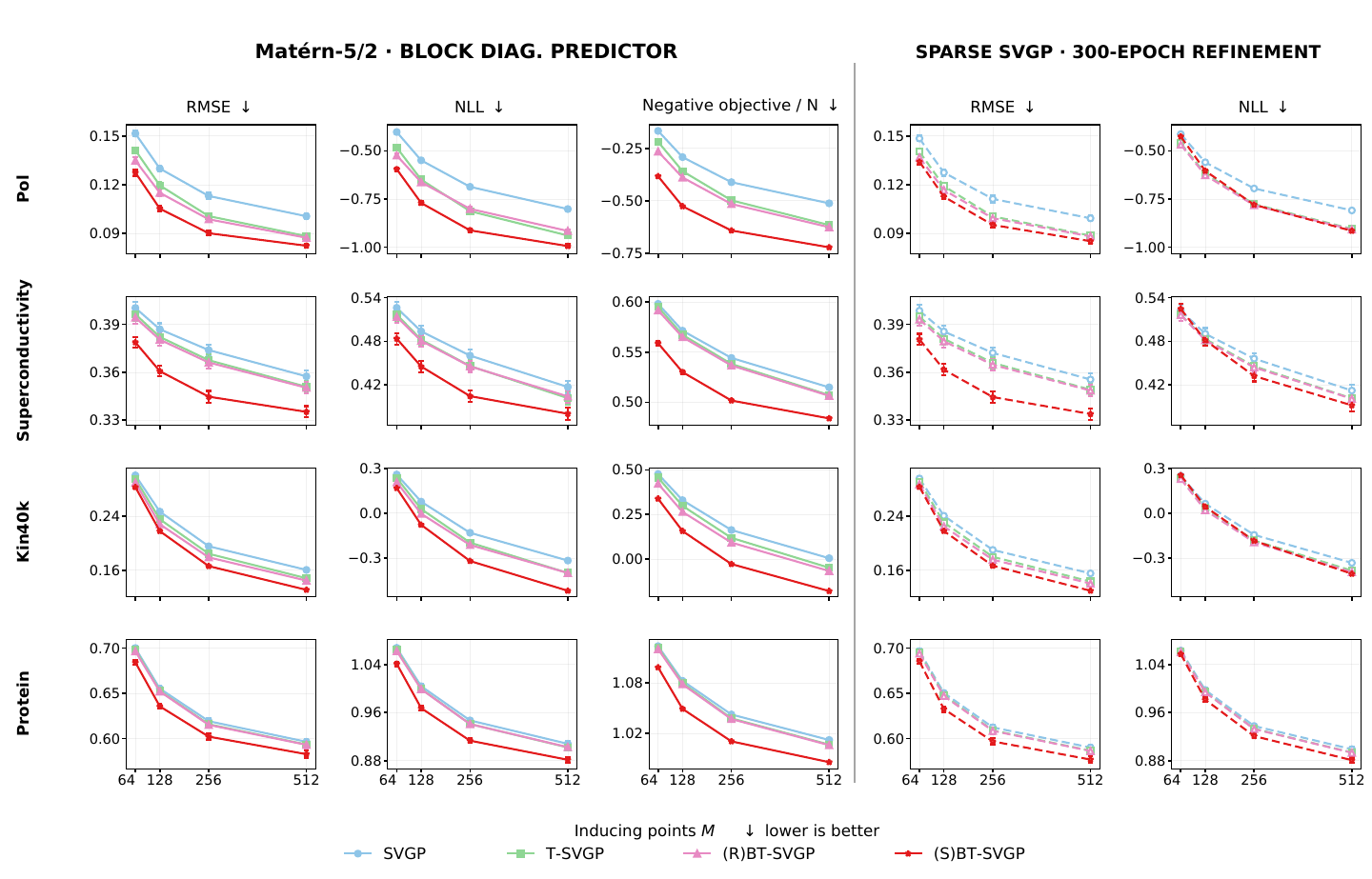}
\caption{Mat\'ern-5/2 SVGP campaign: Pol, Superconductivity, Kin40k, Protein. Left: block-diagonal predictor and negative final objective (solid); right: sparse predictor after the transplant (dashed). Mean and one standard error over ten seeds.}
\label{fig:campaign-matern52-main}
\end{figure}
\end{landscape}

\begin{landscape}
\begin{figure}[p]
\centering
\includegraphics[width=.99\linewidth,height=.83\textheight,keepaspectratio]{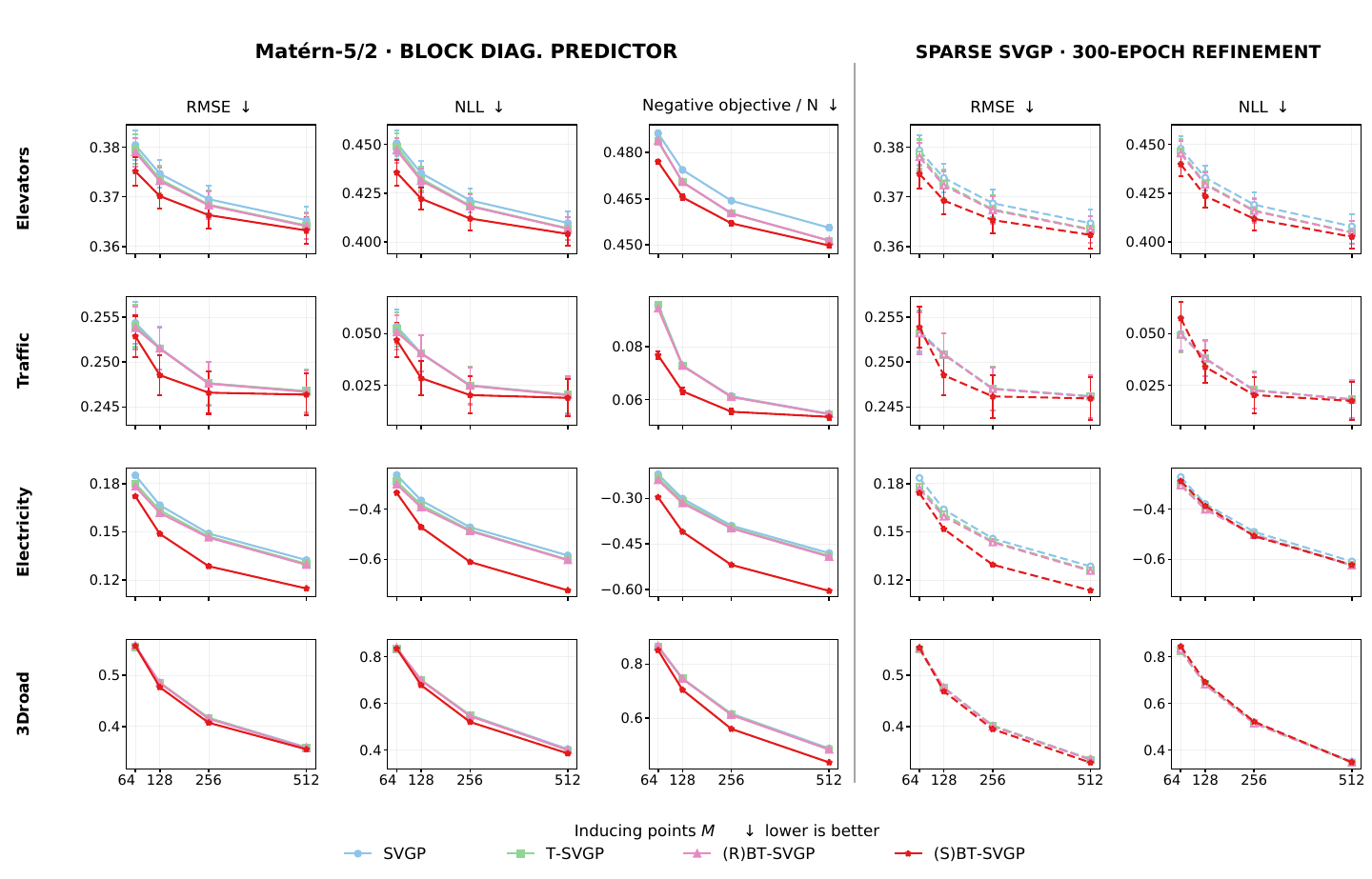}
\caption{Mat\'ern-5/2 SVGP campaign: Elevators, Traffic, Electricity, 3Droad. Left: block-diagonal predictor and negative final objective (solid); right: sparse predictor after the transplant (dashed). Mean and one standard error over ten seeds.}
\label{fig:campaign-matern52-additional}
\end{figure}
\end{landscape}

\begin{landscape}
\begin{figure}[p]
\centering
\includegraphics[width=.99\linewidth,height=.83\textheight,keepaspectratio]{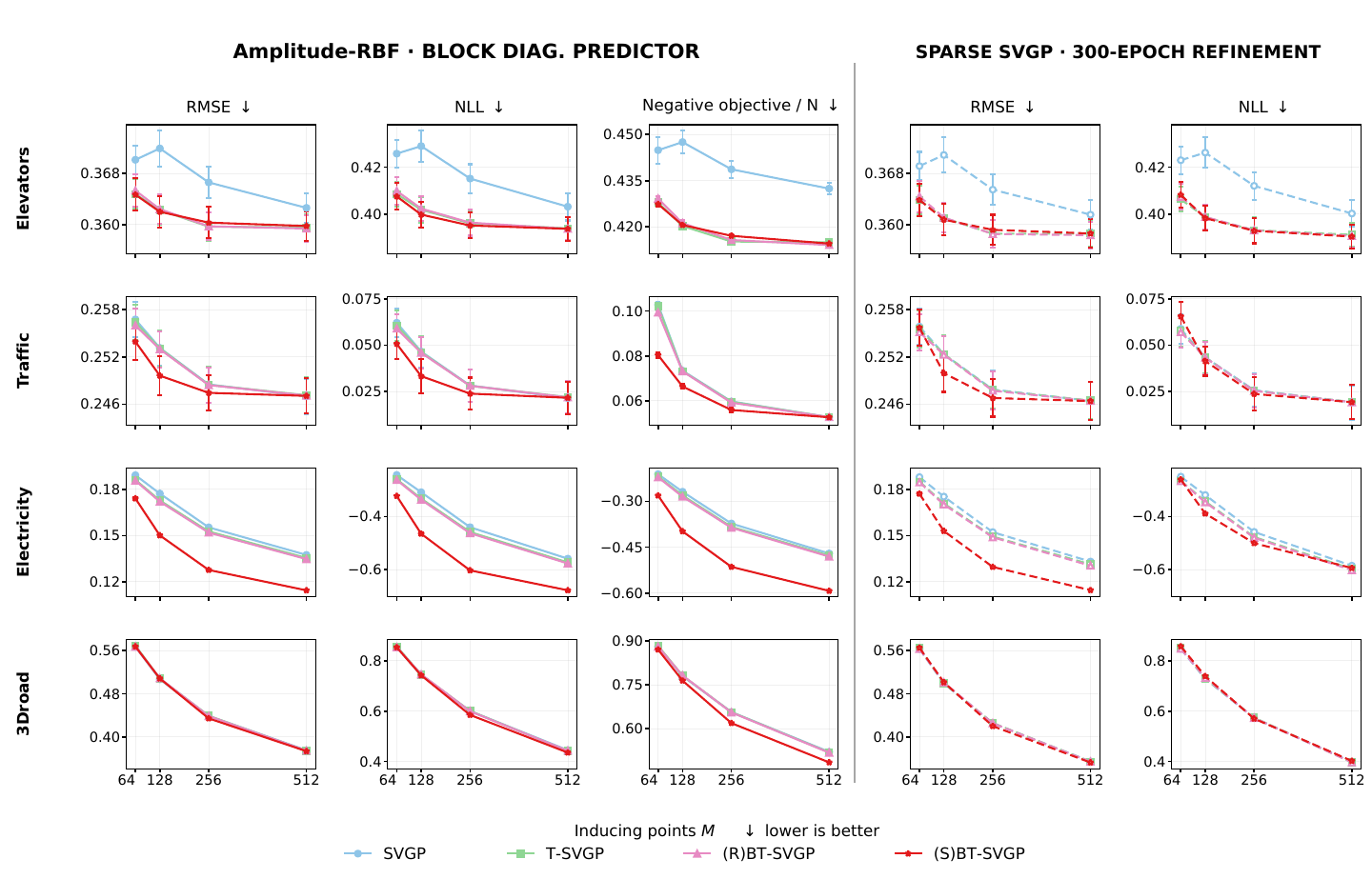}
\caption{Amplitude-RBF SVGP campaign: Elevators, Traffic, Electricity, 3Droad. Left: block-diagonal predictor and negative final objective (solid); right: sparse predictor after the transplant (dashed). Mean and one standard error over ten seeds.}
\label{fig:campaign-amplitude_rbf-additional}
\end{figure}
\end{landscape}

\begin{landscape}
\begin{figure}[p]
\centering
\includegraphics[width=.99\linewidth,height=.83\textheight,keepaspectratio]{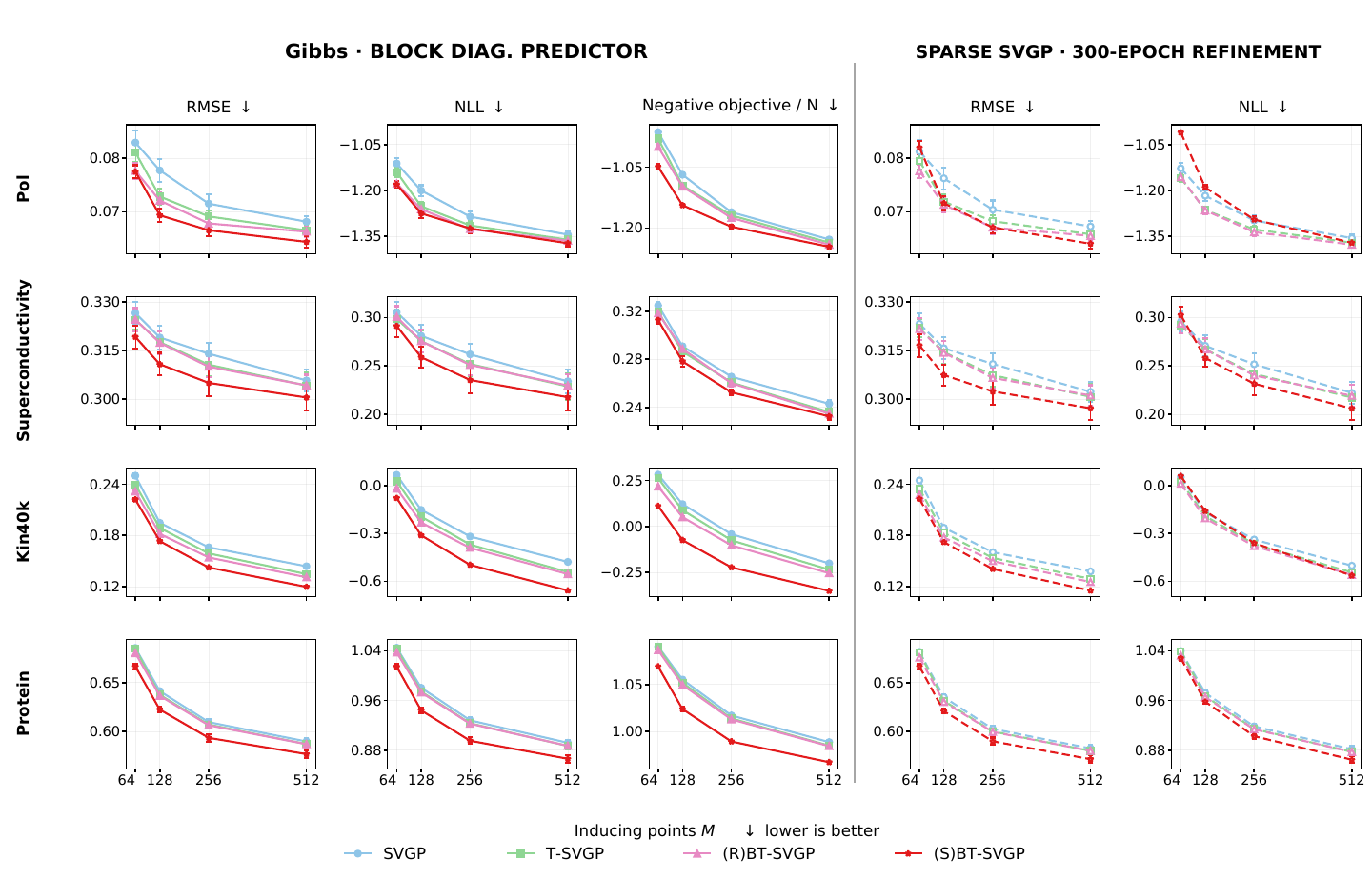}
\caption{Gibbs SVGP campaign: Pol, Superconductivity, Kin40k, Protein. Left: block-diagonal predictor and negative final objective (solid); right: sparse predictor after the transplant (dashed). Mean and one standard error over ten seeds.}
\label{fig:campaign-gibbs-main}
\end{figure}
\end{landscape}

\begin{landscape}
\begin{figure}[p]
\centering
\includegraphics[width=.99\linewidth,height=.83\textheight,keepaspectratio]{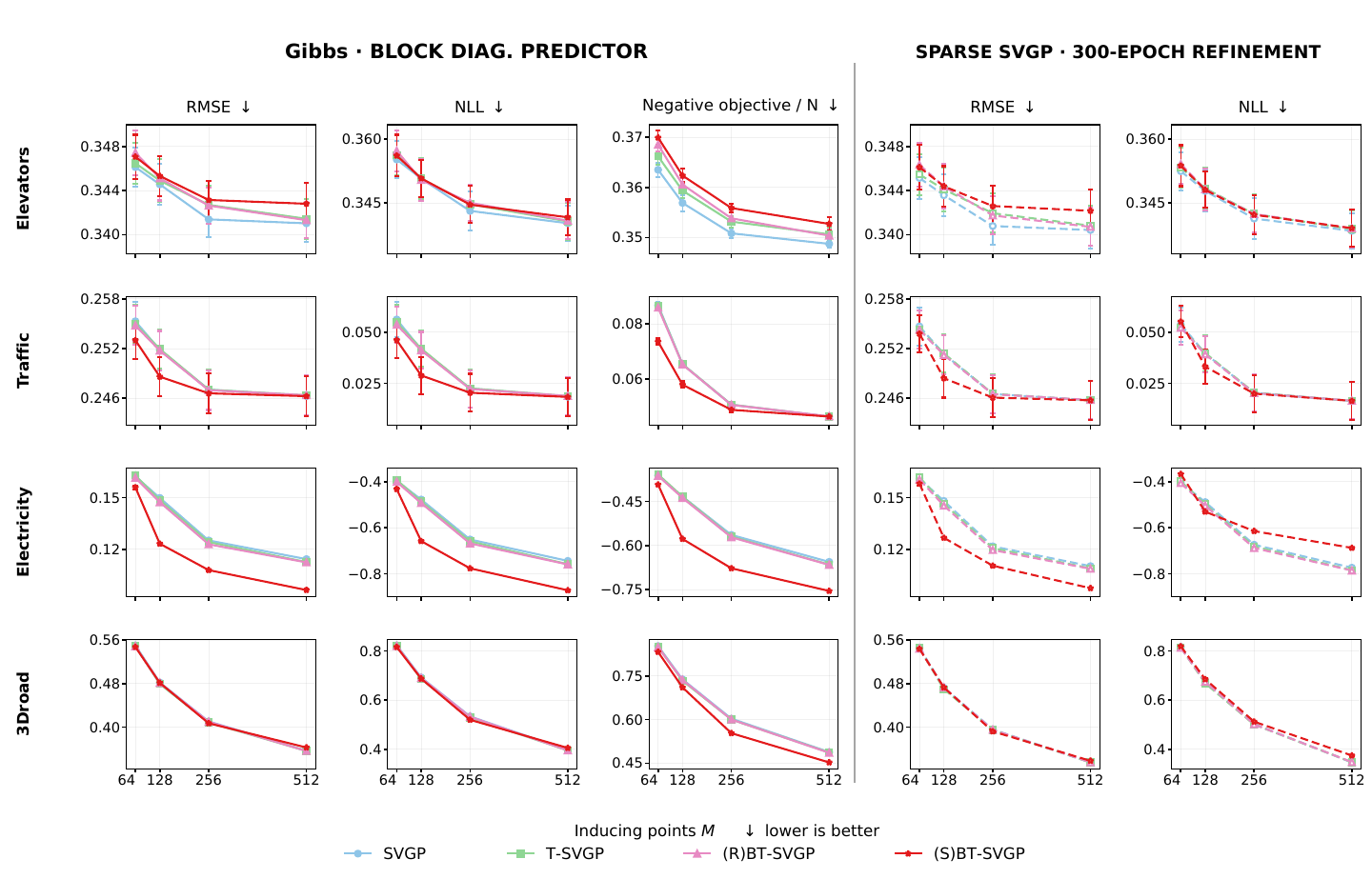}
\caption{Gibbs SVGP campaign: Elevators, Traffic, Electricity, 3Droad. Left: block-diagonal predictor and negative final objective (solid); right: sparse predictor after the transplant (dashed). Mean and one standard error over ten seeds.}
\label{fig:campaign-gibbs-additional}
\end{figure}
\end{landscape}

Figure~\ref{fig:svgp-final} and
Figures~\ref{fig:campaign-rbf-additional}--\ref{fig:campaign-gibbs-additional}
report all 5120 donor runs and their 5120 sparse transplants: four kernels,
eight datasets (Pol, Superconductivity, Kin40k, Protein, Elevators
\citep{torgoElevators}, Traffic \citep{uciTraffic2019}, Electricity
\citep{uciElectricity2015}, 3Droad \citep{uci3droad2013}),
\(M\in\{64,128,256,512\}\), four methods and ten seeds. Elevators, Kin40k,
Protein and 3Droad come from the \texttt{uci\_datasets} package
\citep{evansUciDatasets}. In each figure the left three columns show the
block-diagonal predictor and the negative last-epoch stochastic objective per
datum (solid), and the right two columns the sparse SVGP predictor after the
300-epoch transplant (dashed); matched panels share vertical axes. Means and
one standard error over ten seeds. The objectives are method-specific
stochastic estimates, not a common marginal likelihood.
The solid T-SVGP, (R)BT-SVGP and (S)BT-SVGP curves use the predictor of
Appendix~\ref{app:hard-block-predictor}, each with the prediction partition of
Appendix~\ref{app:prediction-partitions}.

\clearpage

\clearpage

\clearpage

\clearpage

\clearpage

\clearpage

\end{document}